\documentclass[letterpaper]{article}
\usepackage[preprint]{aaai2027}
\usepackage[hyphens]{url}
\usepackage{graphicx}
\usepackage{amsmath,amssymb}
\usepackage{natbib}
\usepackage{caption}
\usepackage{booktabs}
\usepackage{tabularx}
\usepackage{placeins}

\newcommand{\method}{SEER}

\title{SEER: A Self-Grounded Evidence Interface for Controlled Spatial Relation Classification}
\author{Feixiang Liu\textsuperscript{1,2}, Likun Wang\textsuperscript{1,2}, Qiang Qiu\textsuperscript{1}, Hui Xu\textsuperscript{1}, Huawei Shen\textsuperscript{1}, Xueqi Cheng\textsuperscript{1}}
\affiliations{
\textsuperscript{1}State Key Laboratory of AI Safety, Institute of Computing Technology, CAS\\
\textsuperscript{2}University of Chinese Academy of Sciences\\
\{liufeixiang23s@mails.ucas.ac.cn, wanglikun23z@ict.ac.cn, qiuqiang@ict.ac.cn, xuhui@ict.ac.cn, shenhuawei@ict.ac.cn, cxq@ict.ac.cn\}
}

\begin{document}

\maketitle

\begin{abstract}
Spatial relation questions require a model to identify the queried subject and object before comparing their layout. Yet a VLM can recognize both entities and still answer from the wrong instance or an ambiguous global view. We ask whether making query-specific evidence explicit can mitigate this failure and propose SEER (Self-grounded Evidence for Entity-Relation Reasoning), a training-free inference-time evidence interface for frozen VLMs. SEER hides candidate relations during pair localization, constructs a query-specific view with explicit subject/object roles, and retains the full image and sparse box geometry as complementary evidence. For relation-choice protocols with exact inverse support, an optional refinement swaps the entity roles and changes the forward decision only when exactly one visual state obeys the corresponding inverse relation. On an image-disjoint GQA-Train900 test frozen before model scoring, SEER pools to +3.94 [2.17,5.72] over Full; the gain remains positive under label-independent grounding-order counterbalancing and on the 535 rows whose entity names are unique. The unchanged protocol yields +4.35 to +11.79 on all 2,434 filtered EmbSpatial pair-relation questions across three models. Matched controls separate local refocus from role-explicit conditioning. These results establish query-specific evidence construction as the principal intervention, with reciprocal consistency as a smaller protocol-specific refinement.
\end{abstract}

\noindent\textbf{Code:} \url{https://github.com/SouthWinter/SEER}

\section{Introduction}

Vision-language models (VLMs) increasingly answer questions that require reasoning about object relationships, yet spatial relations remain a stubborn failure mode. Benchmarks such as VSR \citep{liu2023vsr}, What'sUp \citep{kamath2023whatsup}, GQA \citep{hudson2019gqa}, ARO \citep{yuksekgonul2023bags}, and recent spatial-intelligence surveys \citep{yu2025sibench} show that models can recognize individual objects while failing to compare their positions. These failures matter because spatial relation reasoning is a core step in visual question answering, embodied instruction following, diagram understanding, and safety-critical visual assistance.

Reasoning-oriented multimodal checkpoints are designed to support extended deliberation, but checkpoint identity alone does not establish correct visual evidence selection. A spatial answer is only as good as its input evidence: attending to the wrong instance, missing a small reference object, or relying on a language prior can corrupt the premise before relation prediction. This is consistent with recent hallucination work: HallusionBench \citep{guan2024hallusionbench} stresses image-context reasoning, and Reallocating Attention \citep{lu2026reallocating} argues that multimodal failures involve an imbalance between perceptual and reasoning stages. We use \emph{evidence alignment} operationally: whether exposing query-specific S/O evidence improves the final decision, not whether predicted boxes exactly recover annotated instances.

\method{} makes query-specific evidence explicit before relation prediction. Target-relation-hidden self-grounding constructs a local S/O-Marker view only for the queried pair: it shows where the entities are and which plays each linguistic role without candidate relations during localization. Unlike generic or detector-generated marks, the evidence comes from a frozen VLM grounder queried for that pair. Full preserves context and threshold-eligible Geometry supplies a sparse layout cue. A fixed forward rule produces an answer; inverse-supported protocols may additionally use reciprocal consistency as a conflict-resolution refinement. No model is trained and Full remains a fallback. The methodological unit is not a new crop, glyph, or box heuristic in isolation, but their leakage-controlled composition into pair-specific evidence with explicit roles, selective geometry, and a deterministic Full fallback.

Our evaluations separate frozen confirmation from development and mechanism analysis. On the same-source, image-disjoint GQA-Train900 nine-relation test frozen before scoring, Qwen3-Instruct/Thinking gain +3.11/+4.78 over Full; the pooled gain is +3.94 [2.17,5.72], and remains +2.56 [0.72,4.39] under label-independent hash counterbalancing. The complete GQA/VG development pools contain 41,544 Qwen3 evaluations, and balanced 900-example sets add Qwen2.5-Instruct and R1-Onevision; final SEER improves all eight model/dataset rows over Full. Under a separately frozen EmbSpatial protocol, Qwen3-Instruct, Qwen3-Thinking, and InternVL3.5 gain +4.35, +5.09, and +11.79 on all 2,434 filtered pair-relation questions. Matched Crop, Self-Marker, corrupted-marker, and Oracle-Marker controls identify local evidence and explicit entity roles as the dominant evidence intervention.

Our contributions are:
\begin{itemize}
    \item We formulate spatial answering as a query-specific evidence-interface problem and introduce target-relation-hidden self-grounding that turns the queried pair into role-explicit visual evidence without VLM training or an external detector.
    \item As an optional inverse-supported refinement, we use exact subject/object exchange to obtain an observable consistency cue for conflicting evidence states, without learned calibration or target labels.
    \item We confirm gains on a nine-relation test frozen before scoring and on EmbSpatial, audit entity-order and referential-ambiguity sensitivity, and isolate local refocus, role-explicit conditioning, and grounding headroom through matched counterfactuals and human audits.
\end{itemize}

\section{Related Work}

\paragraph{Spatial and Compositional Failures.}
Compositional vision-language evaluation has repeatedly shown that models can recognize entities while failing to bind relations, word order, or object roles \citep{antol2015vqa,goyal2017vqa,johnson2017clevr,suhr2019nlvr2,thrush2022winoground,parcalabescu2022valse,yuksekgonul2023bags,hsieh2023sugarcrepe}. Spatial reasoning exposes this gap particularly clearly: prior work tests whether VLMs understand prepositions, relative direction, metric layout, and grounded scene relationships rather than only object presence or language priors \citep{krishna2017visualgenome,hudson2019gqa,liu2023vsr,kamath2023whatsup,rajabi2024gsrbench,chen2024spatialvlm,jia2025omnispatial,yu2025sibench}. We use this literature as the problem basis, not as a dataset contribution: our question is whether failures can be repaired by changing the visual evidence state available at inference time.

\paragraph{Global-Image VLM Interfaces.}
Large-scale image-text pretraining and interleaved multimodal modeling made global-image VLMs broadly useful for recognition, retrieval, captioning, and visual dialogue \citep{radford2021clip,li2022blip,li2023blip2,alayrac2022flamingo,dai2023instructblip,liu2024llava15,peng2024kosmos2}. However, a global image interface leaves instance selection implicit. When a question asks about a particular subject/object pair, the model must internally decide which instances matter, which view to inspect, and which relation cue to compare. \method{} treats this interface as the bottleneck: it does not replace the VLM, but exposes competing evidence states so that the answer need not rely on an unobserved binding step.

\paragraph{Visual Prompting and Grounding Interfaces.}
Open-vocabulary grounding and segmentation systems make region-level visual prompts practical \citep{liu2024groundingdino,kirillov2023sam}, and multimodal LLMs have been extended with grounded reference interfaces that connect language to regions \citep{peng2024kosmos2,chen2023shikra,you2024ferret}. Region-aware and spatially trained models such as SpatialRGPT and SpatialVLM improve grounded spatial reasoning with region proposals, depth, or spatial training data \citep{cheng2024spatialrgpt,chen2024spatialvlm}. These systems define stronger model- or tool-augmented settings; our target is instead an inference-time interface for an unchanged VLM. Training-free visual prompting takes a complementary path: Set-of-Mark overlays visible marks so that a model can refer to regions using language-like symbols \citep{yang2023som}, VP-Bench systematically tests whether MLLMs can perceive and use such prompts \citep{xu2026vpbench}, and Graph-of-Mark adds graph-connected region cues for spatial relations \citep{frisoni2026graphofmark}. \method{} follows the same broad intuition that entities must become visually speakable, but changes the evidence protocol: target-relation-hidden self-grounding replaces external detection, the visual intervention is restricted to the queried pair, and geometry is admitted only when that pair's self-grounded layout is unambiguous. These choices target pair-specific role conditioning rather than dense scene description; the supplement includes a matched diagnostic using the released Graph-of-Mark API.

\paragraph{Reasoning, Hallucination, and Evidence Reliability.}
Chain-of-thought and reasoning-oriented multimodal models motivate more deliberate inference \citep{wei2022cot}; hallucination work shows the complementary risk that fluent reasoning can drift away from image evidence \citep{li2023pope,guan2024hallusionbench}. MM-R3 measures consistency under question rephrasing, image restyling, and context changes and trains an adapter to reduce inconsistency \citep{chou2025mmr3}; test-time consistency likewise aligns generic semantics-preserving variants through gradient-based adaptation \citep{chou2026testtime}. SEER instead changes the visual evidence interface without adaptation and uses subject/object exchange as a task-specific relation equivariance only when an exact inverse exists. Other intervention methods modify decoding, attention, or internal mechanisms. Reallocating Attention, for example, rescales heads/layers to reduce MLRM hallucination \citep{lu2026reallocating}. Our work instead externalizes subject/object evidence and uses this transformation only as a protocol-specific conflict cue.

\begin{figure*}[!t]
\centering
\includegraphics[width=\textwidth]{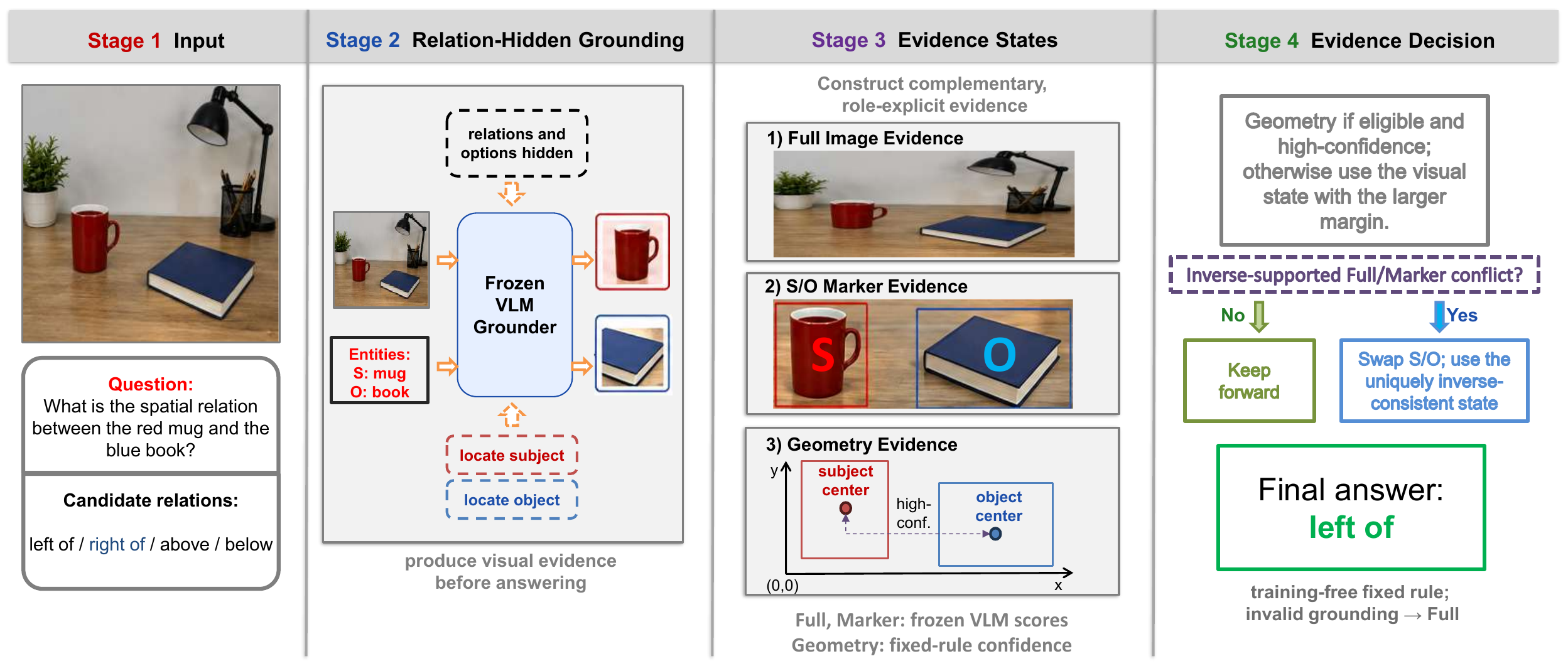}
\caption{\method{} overview. Relation-hidden grounding builds Full, S/O-Marker, and Geometry evidence. The training-free rule uses eligible Geometry or the larger-margin visual state, resolves eligible conflicts by inverse-consistent S/O exchange, and falls back to Full after invalid grounding.}
\label{fig:method}
\end{figure*}

\section{Problem Setup}

We study spatial relation decisions. For relation multiple choice, we write an example as $x_i=(I_i,s_i,o_i,\mathcal{A}_i)$, where $I_i$ is the image, $s_i$ and $o_i$ are the subject and object phrases, and $\mathcal{A}_i$ is the candidate answer set. For caption-pair evaluation, $\mathcal{A}_i$ contains the two candidate captions whose subject/object relation differs. The goal is to select the visually correct answer for the specified subject/object pair. We operationalize evidence alignment as the effect of exposing query-conditioned S/O evidence on that final decision: a model may know spatial vocabulary, or even localize objects in isolation, yet still answer from a text prior, attend to the wrong instance, or rely on an ambiguous global view. Exact reference localization is audited separately rather than inferred from answer accuracy.

The key constraint is that self-grounding must not receive the gold relation. During grounding, the model sees the image, the subject/object names, and generic pair-localization wording; answer options and relation labels are hidden. This prevents the grounder from using the answer relation to choose convenient boxes. For ARO caption-pair examples, the relation phrase is visible in both candidate captions; we therefore report ARO separately as an answer-blind caption-pair transfer diagnostic and keep GQA/VG as the cleaner target-relation-hidden relation-classification setting.

\section{Method}

\method{} is an inference-time evidence interface for a frozen VLM $M$ and frozen VLM grounder $G$. By default $G=M$; transfer experiments may fix $G$ while changing $M$. The procedure trains no model and uses no external detector. As illustrated in Figure~\ref{fig:method}, SEER first localizes the queried pair without access to candidate relations, converts that localization into role-explicit evidence, and obtains state-wise predictions from the same $M$. It then makes a fixed forward decision. When the answer protocol supports an exact inverse transformation, an optional reciprocal step may refine conflicts by exchanging the subject and object; Marker and Forward remain separately reported to expose the contribution of evidence construction.

The design has three constraints. First, localization is target-relation-hidden, so boxes cannot be chosen after seeing the target relation. Second, the same frozen VLM $M$ reads every visual-linguistic state in a given experiment; changing $G$ changes only the evidence source. Third, the decision reuses state predictions and a known inverse-relation map rather than fitting a selector. Evidence construction supplies the visual intervention; optional reciprocal consistency refinement supplies a protocol-specific transformation cue.

\subsection{Target-Relation-Hidden Self-Grounding}

Given the answer-removed localization input $x_i\setminus\mathcal{A}_i=(I_i,s_i,o_i)$, grounder $G$ outputs normalized boxes $b_{s_i},b_{o_i}\in[0,1]^4$. Its prompt names the two entities but excludes $\mathcal{A}_i$, the candidate relation, and any directional hint. We use subject-first as the pre-specified realization because the queried relation is the ordered function $r(s_i,o_i)$; this semantic convention is deterministic but is not claimed to remove mention-order bias. A sample-ID-hash counterbalancing audit is reported separately. When $G=M$, this is same-model self-grounding; when a shared $G$ is used, we state it explicitly.

We validate each box before using it. Boxes must have positive area, lie inside the image after clipping, and not collapse to degenerate coordinates. If either box is invalid, the example falls back to full-image evidence so that every method is evaluated on the same denominator. This fallback is important for interpretation: improvements cannot come from silently dropping difficult localization cases.

\subsection{Constructing Complementary Evidence Types}

The evidence types are deliberately complementary. Full image preserves context, marker view exposes role-explicit local conditioning, and geometry tests whether the box layout alone is enough. This redundancy is useful because a state can be informative on one example and misleading on another.

\paragraph{Full image.}
The unmodified image and original question. This preserves global context and is the safest fallback when grounding is uncertain, when the relevant relation depends on scene context, or when local cropping removes useful cues.

\paragraph{Self-grounded marker view.}
We crop a padded union of the subject/object boxes and draw a red box labeled S around the subject and a blue box labeled O around the object. Padding preserves nearby context while reducing unrelated distractors. The prompt states that S is the subject and O is the object and asks the model to answer about the S-to-O pair. This state combines three effects: local refocus, explicit role assignment, and a prompt-level interface for relation reading. Treating it as a separate state lets the experiments distinguish role-explicit S/O conditioning from cropping alone.

\paragraph{Threshold-eligible geometry.}
We derive a geometric candidate from self-grounded box centers and extents. In the target-complete protocol, Geometry is eligible for left/right via signed horizontal displacement and for inside/contains via asymmetric box coverage. It abstains for the above/below/on family because ``on'' cannot be distinguished reliably from vertical displacement alone, for front/behind because 2D boxes do not encode depth, and whenever its normalized layout score is below threshold. Geometry is therefore a sparse thresholded source available only on eligible rows, not a third VLM prediction, a calibrated confidence estimate, or a universal replacement for visual reasoning.

\subsection{Evidence States and Predictions}

For each example $x_i$, \method{} defines three complementary evidence types for the same subject/object pair. Full and Marker are visual-linguistic states read by $M$; Geometry is an optional deterministic state that carries a box-derived candidate and layout score and does not invoke $M$. Let $\mathcal{V}_i\subseteq\{\mathrm{full},\mathrm{marker}\}$ denote the available visual states and $E_i^{(t)}$ the concrete input shown to $M$ for $t\in\mathcal{V}_i$. Full is always available, Marker requires valid grounding, and Geometry may abstain; the procedure therefore does not compare three VLM predictions on every row.

The baseline applies $M$ only to Full, whereas \method{} applies the same frozen $M$ to each available visual state. For an answer option $a\in\mathcal{A}_i$ and $t\in\mathcal{V}_i$, let $\ell_i^{(t)}(a)$ be the closed-set log-probability assigned by $M$ under $E_i^{(t)}$. The visual-state prediction $\hat{a}_i^{(t)}$ and option-score gap $m_i^{(t)}$ are
\begin{equation}
\label{eq:state-prediction}
\begin{array}{rcl}
\hat{a}_i^{(t)} &=& \arg\max_{a\in\mathcal{A}_i} \ell_i^{(t)}(a),\\
m_i^{(t)} &=& \ell_i^{(t)}(\hat{a}_i^{(t)})
-\max_{a\in\mathcal{A}_i,\,a\neq \hat{a}_i^{(t)}} \ell_i^{(t)}(a).
\end{array}
\end{equation}
The option-score gap is the difference between the top option and the strongest alternative. It is an uncalibrated ordering cue rather than a probability of state correctness, and it does not create a new answer candidate.

\subsection{Training-Free Evidence Decision}

The decision stage does not predict a new relation; it returns one of the state predictions defined above. It first obtains a forward candidate and then checks reciprocal consistency only for eligible visual-state conflicts. Both operations are deterministic and use no supervision or learned selector.

\noindent\textbf{Forward decision.} Let $\hat{a}_i^{(\mathrm{geom})}$ be the box-derived candidate, $c_i^{(\mathrm{geom})}$ its normalized layout score, and $z_i^{(\mathrm{geom})}$ indicate that the candidate maps to a supported relation. Geometry takes priority only when it is eligible and $c_i^{(\mathrm{geom})}\ge\tau$. Otherwise, the visual state with the larger option-score gap provides a fallback forward source; exact ties favor Full. This source is
\begin{equation}
\label{eq:forward-source}
\tilde{t}_i=\begin{cases}
\mathrm{geom}, & z_i^{(\mathrm{geom})}=1\ \wedge\ c_i^{(\mathrm{geom})}\ge\tau,\\
\mathrm{marker}, & \mathrm{marker}\in\mathcal{V}_i\ \wedge\ m_i^{(\mathrm{marker})}>m_i^{(\mathrm{full})},\\
\mathrm{full}, & \text{otherwise,}
\end{cases}
\end{equation}
The gap is only an ordering cue, not a calibrated probability that a state is correct. We use $\tau=0.5$ for every model and dataset, and invalid grounding returns Full.

\noindent\textbf{Reciprocal consistency refinement.} Let $r(a)$ map an option to its relation and $\iota(r)$ denote the exact inverse, such as left-of$\leftrightarrow$right-of or inside$\leftrightarrow$contains. This refinement is used only in protocols whose candidate relations have defined inverse mappings and that guarantee an inverse candidate for the correct relation; eligibility is fixed by dataset construction, not by model outputs. On a row, Geometry must abstain and Full and Marker must disagree. We exchange $s_i$ and $o_i$ while preserving the image, option strings, and order; for Marker, the visible S/O roles are also exchanged. If $\hat a_{i,\rho}^{(t)}$ is state $t$'s prediction for this reciprocal query, its consistency indicator is $q_i^{(t)}=\mathbb 1[r(\hat a_{i,\rho}^{(t)})=\iota(r(\hat a_i^{(t)}))]$, which is zero if the predicted relation's inverse is absent. The final source is
\begin{equation}
\label{eq:reciprocal-decision}
t_i^*=\begin{cases}
\mathrm{full}, & q_i^{(\mathrm{full})}=1,\ q_i^{(\mathrm{marker})}=0,\\
\mathrm{marker}, & q_i^{(\mathrm{marker})}=1,\ q_i^{(\mathrm{full})}=0,\\
\tilde t_i, & \text{otherwise.}
\end{cases}
\end{equation}
SEER returns $\hat a_i^{(t_i^*)}$. Thus reciprocity does not average scores or certify correctness by itself: it changes the forward source only when one state, but not the other, respects the required relation transformation. Protocols without inverse support retain the forward decision. The supplement reports unconditional score fusion, equal-call reciprocal alternatives, trigger rates, and binary-task boundaries.

\section{Experimental Setup}

\paragraph{Datasets.}
Our target-complete development suites contain 11,070 GQA \citep{hudson2019gqa} and 9,702 Visual Genome (VG) examples over nine relations. Targets are balanced within four option families: left/right, above/below/on, front/behind, and inside/contains. Every displayed relation occurs as a target; near/far is excluded because the source annotations contain almost no reliable far targets. We also sample 900 examples per dataset, 100 per relation, for balanced four-model development analysis. After fixing the method, we froze a separate GQA-Train900 test before model scoring: it contains 100 examples for each relation, uses 900 unique images, and has zero image-ID overlap with either development pool. We treat these constructed sets as controlled diagnostics; the development source pairs and a targeted 120-row frozen sample receive independent semantic audits. From the official EmbSpatial-Bench test \citep{du2024embspatial}, we retain all 2,434 rows with exactly two queried objects and a left/right/above/under target, spanning 276 scenes. We preserve its questions, options, order, and labels; official boxes are withheld from SEER and used only for post-hoc grounding IoU. The complete public VSR random test \citep{liu2023vsr} and all 1,860 inverse-predicate SpatialSense test judgments \citep{yang2019spatialsense} probe binary transfer. COCO-Geom is a 400-example box-derived mechanism diagnostic \citep{lin2014coco}; ARO VG-Relation provides an external caption-pair test \citep{yuksekgonul2023bags}. Construction and audit details are in the supplement.

\paragraph{Models and scoring.}
The complete-pool study uses Qwen3-VL-8B-Instruct and Qwen3-VL-8B-Thinking \citep{bai2025qwen3vl}. Balanced evaluation adds Qwen2.5-VL-7B-Instruct \citep{bai2025qwen25vl} and R1-Onevision-7B \citep{yang2025r1onevision}. Qwen3-I/T use same-model grounding ($G=M$). For a controlled shared-grounder transfer setting, Qwen2.5-I and R1-OV use a fixed Qwen3-I grounder; this also avoids R1-OV's unreliable box output through its tested interface. In every row, $M$ remains the named evaluated VLM. Architecture-diverse replications use InternVL3.5-8B-HF with $G=M$ on GQA and EmbSpatial \citep{wang2025internvl35}; ARO additionally includes Qwen3-2B and LLaVA-1.5-7B \citep{liu2024llava15}. Tables use the abbreviations above. Relation-MC and caption-pair tasks use length-normalized option log-probabilities with identical options and order across visual states.

\paragraph{Baselines and metrics.}
The primary baseline applies the same VLM to the original image. Crop isolates local refocus, Marker uses only the role-explicit S/O-marker state, Forward is the pre-refinement output of Equation~\ref{eq:forward-source}, and \method{} applies Equation~\ref{eq:reciprocal-decision} whenever eligible. Accuracy is computed on every example; paired wrong-to-right/right-to-wrong counts and exact McNemar tests compare methods on identical rows. The supplement reports prompt-only SOP, Text-BBox, same-image prompt ensembles, external-grounder controls, the released Graph-of-Mark high-level API, unconditional score fusion, and equal-call reciprocal alternatives.

\paragraph{Decision protocols and statistics.}
All methods are evaluated on identical examples. The decision rule and global threshold $\tau=0.5$ are fixed across models and datasets, with no dataset- or model-specific fitting. GQA/VG validation data served as method-development suites. We then applied the unchanged relation-MC protocol to the image-disjoint GQA-Train900 split and frozen EmbSpatial evaluation; VSR reused the frozen evidence construction and threshold under its native binary scoring. The complete 1,860-example SpatialSense inverse-predicate subset was separately frozen before scoring; a later relation-MC adaptation is labeled post-hoc in the supplement. The full chronology is documented there. In a post-hoc development-pool audit, two independent annotators inspected 240 stratified predicted groundings without model/dataset/IoU metadata and 270 relation-balanced source pairs without gold labels. Both university-peer annotators volunteered without compensation, gave informed consent, and provided no identifying or sensitive information. Because one image can yield several relation tuples, primary uncertainty uses a paired nonparametric bootstrap that resamples image IDs or scene IDs (20,000 repetitions). For pooled GQA/VG intervals, identical numeric image IDs form one cluster because GQA is derived from VG. GQA-Train900 resamples its 900 unique images, clustering both model outcomes by image when pooled. We report 95\% intervals for gains; pooled rows remain descriptive and all gains are computed from exact counts before rounding. The accompanying anonymous artifact provides frozen IDs and checksums, construction and scoring code, prompts, per-option scores, and analysis scripts.

\section{Frozen In-Domain Confirmation and Order Sensitivity}

Table~\ref{tab:frozen_gqa_train} evaluates GQA-Train900, whose IDs and checksum were fixed before model scoring. Pre-specified subject-first SEER pools to +3.94 [2.17,5.72] over Full; complete object-first and label-independent Hash controls retain +2.22 [0.33,4.11] and +2.56 [0.72,4.39]. A more expensive two-order Marker score mean gives +3.00 [0.89,5.11]. Every pooled order control is positive, although relation-level effects remain heterogeneous. Subject-first gains +1.22 over both Marker and Forward, identifying evidence construction as the larger intervention.

\begin{table}[!ht]
\centering
\small
\setlength{\tabcolsep}{1.2pt}
\begin{tabular*}{\columnwidth}{@{\extracolsep{\fill}}lrrrrrr@{}}
\toprule
Setting & Full & Marker & Fwd. & \method{} & Gain & $\mathrm{CI}_{img}$ \\
\midrule
S-first/I & 75.33 & 76.11 & 77.44 & \textbf{78.44} & +3.11 & [1.00,5.22] \\
S-first/T & 70.56 & 75.22 & 73.89 & \textbf{75.33} & +4.78 & [2.56,7.11] \\
\midrule
S-first/All & 72.94 & 75.67 & 75.67 & \textbf{76.89} & +3.94 & [2.17,5.72] \\
\midrule
Object/All & 72.94 & 74.33 & 73.56 & \textbf{75.17} & +2.22 & [0.33,4.11] \\
Hash/All & 72.94 & 74.72 & 74.17 & \textbf{75.50} & +2.56 & [0.72,4.39] \\
\bottomrule
\end{tabular*}
\caption{Frozen image-disjoint GQA-Train900 results (900 unique images; nine relations, 100 each). S-first is pre-specified; Object and Hash are post-hoc order controls. I/T denote Qwen3-I/T. Gain and intervals compare final SEER with Full.}
\label{tab:frozen_gqa_train}
\end{table}

Under the pre-specified protocol, reciprocal refinement triggers on 10.61\% of model--example pairs and makes 40 fixes versus 18 breaks relative to Forward, a pooled +1.22 [0.39,2.06]. Under Hash, final SEER versus Full gives per-model $p=.343/.00156$; relation-wise results are in the supplement.

\section{Development Results on GQA and VG}

Table \ref{tab:main_gqa_vg} gives results on the complete constructed pools. Final SEER improves Qwen3-I from 72.62 to 75.65 (+3.03) and Qwen3-T from 68.59 to 73.54 (+4.95). All four model/dataset image-cluster intervals against Full exclude zero. Marker supplies the initial +2.18-point aggregate gain, Forward adds +0.45, and reciprocal consistency refinement adds another +1.36 [1.17,1.54] over Forward. The final pooled gain is +3.99 [3.58,4.40] over Full.

\begin{table}[!ht]
\centering
\small
\setlength{\tabcolsep}{1.2pt}
\begin{tabular*}{\columnwidth}{@{\extracolsep{\fill}}llrrrrr@{}}
\toprule
Data & Model & Full & Marker & Fwd. & \method{} & Gain \\
\midrule
GQA & Qwen3-I & 72.40 & 74.03 & 74.27 & \textbf{75.82} & +3.41 \\
GQA & Qwen3-T & 67.10 & 71.20 & 71.08 & \textbf{72.99} & +5.89 \\
VG & Qwen3-I & 72.87 & 73.76 & 74.60 & \textbf{75.47} & +2.60 \\
VG & Qwen3-T & 70.28 & 72.20 & 73.15 & \textbf{74.16} & +3.88 \\
\midrule
All & Qwen3-I & 72.62 & 73.90 & 74.43 & \textbf{75.65} & +3.03 \\
All & Qwen3-T & 68.59 & 71.67 & 72.05 & \textbf{73.54} & +4.95 \\
\bottomrule
\end{tabular*}
\caption{Complete development results on the constructed target-complete pools with same-model grounding (GQA: 11,070; VG: 9,702). Gain compares final SEER with Full.}
\label{tab:main_gqa_vg}
\end{table}

Table \ref{tab:balanced_models} broadens model coverage on relation-balanced subsets. Under subject-first grounding, final SEER improves all eight rows over Full, with a pooled +4.10 points; reciprocal consistency refinement adds +1.21 [0.72,1.70] over Forward. A post-hoc label-independent Hash audit also improves all eight rows and pools to +4.26 [3.27,5.25]. Seven row-wise intervals exclude zero; R1-OV/VG gives +1.89 [$-0.11$,3.92]. Qwen2.5-I and R1-OV use a fixed Qwen3-I grounder, so their rows test interface transfer rather than same-model self-grounding.

\begin{table}[!ht]
\centering
\small
\setlength{\tabcolsep}{1.5pt}
\begin{tabular*}{\columnwidth}{@{\extracolsep{\fill}}llrrrr@{}}
\toprule
Data & Model & Full & S-first & Hash & Gain$_H$ \\
\midrule
GQA & Qwen3-I & 75.00 & 78.67 & \textbf{79.00} & +4.00 \\
GQA & Qwen3-T & 69.11 & \textbf{75.67} & 74.89 & +5.78 \\
GQA & Qwen2.5-I$^\dagger$ & 73.11 & \textbf{78.22} & 77.56 & +4.44 \\
GQA & R1-OV$^\dagger$ & 65.78 & \textbf{70.89} & 70.22 & +4.44 \\
VG & Qwen3-I & 70.33 & 72.67 & \textbf{73.22} & +2.89 \\
VG & Qwen3-T & 66.67 & 71.56 & \textbf{72.33} & +5.67 \\
VG & Qwen2.5-I$^\dagger$ & 68.89 & 72.56 & \textbf{73.89} & +5.00 \\
VG & R1-OV$^\dagger$ & 62.67 & 64.11 & \textbf{64.56} & +1.89 \\
\midrule
All & Overall & 68.94 & 73.04 & \textbf{73.21} & +4.26 \\
\bottomrule
\end{tabular*}
\caption{Balanced cross-model development results (900 examples per dataset; 100 per relation). S-first and Hash report final SEER under the pre-specified and sample-ID-hash grounding orders. Hash assignment uses neither labels nor model outputs; Gain$_H$ compares Hash with Full. $^\dagger$ Grounder: Qwen3-I.}
\label{tab:balanced_models}
\end{table}

As an architecture-diverse same-model replication, InternVL3.5 on GQA900 improves from 61.89 on Full to 63.89 on Marker, 68.78 with Forward, and 70.78 with final SEER (+8.89). We report this single-dataset replication separately rather than alter the symmetric four-model/two-dataset aggregate.

\FloatBarrier

\subsection{Option-Free Sanity Check}

Closed-set scoring controls answer-format variability, so we additionally run a small option-free check with Qwen3-I, directly generating a relation phrase for 20 examples per relation and dataset. The option-free adaptation gains +6.67 on GQA and +5.00 on VG (+5.83 pooled) with 97.50\% parseability; image-cluster intervals are [+2.26,+11.17] and [$-0.56$,+10.61]. This shows feasibility without answer options rather than broad open-generation validity; dataset results and the decision protocol are in the supplement.

\section{Mechanism Evidence}

\paragraph{Controlled geometry.}
COCO-Geom is a controlled mechanism diagnostic derived from COCO instance geometry. Its hard cases include overlap, small margins, diagonal distractors, and non-left/right relations. Self-grounded boxes plus geometry improve Qwen3-T, Qwen3-I, R1-OV, and Qwen2.5-I by +27.75, +21.75, +41.75, and +29.75 points, respectively (full table in the supplement). Because its labels are geometry-derived, COCO-Geom isolates the intended mechanism: query-specific localization recovers spatial labels when box geometry is the relevant decision signal.
Removing Geometry from Forward costs 0.51 [0.40,0.63] points on development and 0.22 [$-0.17$,0.67] when frozen; it is a modest component and never routes EmbSpatial rows.

\paragraph{Matched evidence-state ablation.}

Table \ref{tab:mechanism_ablation} separates local refocus from role-explicit conditioning on the balanced target-complete sets. Crop uses the target-relation-hidden self-grounded union without marks; Self-Marker adds S/O marks using exactly the same boxes. Oracle-Marker replaces self-grounding with source annotation boxes and is an analysis-only upper bound. Swapped and Random hold the self-grounded crop fixed but exchange S/O identities or move same-size markers within the crop. Full$\rightarrow$Crop and Crop$\rightarrow$Self-Marker add +1.25 and +1.64 points overall; the latter has an image-cluster 95\% interval of [+0.25,+3.05]. Oracle-Marker adds another +6.17, whereas Swapped and Random remove 1.67 and 2.50 points from Self-Marker. Thus the gain depends on both local evidence and semantically correct S/O role assignment within the rendered view, while annotation boxes show substantial localization headroom.

\begin{table}[!ht]
\centering
\small
\setlength{\tabcolsep}{1.1pt}
\begin{tabular*}{\columnwidth}{@{\extracolsep{\fill}}lrrrrrr@{}}
\toprule
Model & Full & Crop & Self-M & Oracle-M & Swap & Random \\
\midrule
Qwen3-I & 72.67 & 73.72 & 74.39 & 80.61 & 73.78 & 72.06 \\
Qwen3-T & 67.89 & 69.33 & 71.94 & 78.06 & 69.22 & 69.28 \\
\midrule
All & 70.28 & 71.53 & 73.17 & 79.33 & 71.50 & 70.67 \\
\bottomrule
\end{tabular*}
\caption{Matched target-complete mechanism controls. Oracle-M is an annotation-box upper bound; Swap and Random corrupt S/O role assignment while retaining the self-grounded crop.}
\label{tab:mechanism_ablation}
\end{table}

The complete-pool Qwen3-I expansion sharpens this result under the natural relation mix: Full, Crop, Self-Marker, Forward, and final SEER obtain 72.62, 74.10, 73.90, 74.43, and 75.65, respectively. Marker is slightly below Crop overall because it helps GQA but regresses on VG; reciprocal consistency refinement then adds +1.23 over Forward. Oracle-Marker reaches 80.45, leaving +6.55 points of localization headroom over Self-Marker. Together, the matched controls establish local refocus and role-explicit conditioning as the principal evidence intervention, with relation-aware consistency refinement selecting against some misleading views.

Equal-call reciprocal alternatives and unconditional Full/Marker fusion are reported in the supplement. Across the two frozen GQA and three frozen EmbSpatial rows, the pre-specified rule is always within 0.34 points of the best equal-call alternative, and no alternative dominates across models; we therefore retain it rather than select a rule after observing test labels.

\section{External Transfer and Boundaries}

\paragraph{ARO VG-Relation.}
On our balanced 600-example subset across seven models, Full, Box, and S/O-Marker average 73.14, 77.67, and 80.81; Marker improves six models (+7.67 mean), transferring beyond relation multiple choice. On the complete 23,937-example official split, relation-region crops raise Qwen3-I from 82.51 to 87.08, supporting the localization diagnosis but remaining separate from the self-grounded \method{} claim. Because both captions expose relation words, ARO is answer-blind rather than target-relation-hidden; supplementary crop--box--marker ablations attribute its gain mainly to explicit S/O evidence.

\paragraph{Frozen EmbSpatial confirmation.}
Under the frozen four-direction EmbSpatial protocol, Qwen3-I/T and a later unchanged-protocol InternVL3.5 replication improve over Full by +4.35/+5.09/+11.79, with scene-cluster intervals [+3.12,+5.68], [+3.81,+6.47], and [+10.14,+13.36]. Relative to Marker, final SEER changes by +0.90 [0.07,1.74], +0.78 [$-0.08$,1.60], and $-0.53$ [$-1.63$,0.58]. Reciprocal refinement adds 0.00/+0.25/+0.25 over Forward. Evidence construction transfers strongly; the additional decision benefit is smaller and model-dependent.

\paragraph{Binary protocol boundary.}
On 2,195 VSR random-test judgments, Full/Marker/Forward/SEER obtain 77.86/78.00/78.72/79.04. Final SEER is +1.18 [+0.13,+2.24] over Full, whereas its +0.32 [$-0.14$,+0.81] increment over Forward is not statistically resolved. On 1,860 frozen official SpatialSense annotations, Qwen3-I Full/Forward/SEER obtain 67.69/67.53/67.63, and Qwen3-T obtains 63.82/63.66/63.12 with a fixed Qwen3-I grounder. The interface therefore does not transfer uniformly to binary truth judgment: reciprocal relation consistency is useful for controlled relation choice but is not a general truth-reliability certificate.

\section{Analysis}
\label{sec:analysis}

\paragraph{Evidence construction and its active intervention.}
Across both complete-pool models, Marker improves the descriptive aggregate by +2.18, Forward by +2.63, and final SEER by +3.99. Table~\ref{tab:mechanism_ablation} separates local refocus from role-explicit conditioning, while the oracle gap quantifies localization headroom. Frozen prompt/mark controls and the ARO crop--box--marker progression further show that the active intervention is making query-specific locations and roles jointly inspectable, rather than adding instructions or a particular glyph.

\paragraph{What reciprocal consistency refinement adds.}
The evidence states fail differently: Marker corrects 2,377 complete-pool examples where Full is wrong, while Full preserves 1,436 answers that Marker would break. Raw option-score gaps weakly predict which state is correct; fixed z-score, robust, and percentile normalizations do not improve consistently, and unconditional fusion changes Marker by at most +0.05 points. An oracle over the two visual states is +7.85. Only 4,267 of 41,544 pairs trigger reciprocal consistency refinement, which adds +1.36 [1.17,1.54] over Forward; the frozen test gives +1.22 [0.39,2.06]. The cue is not a visual-correctness certificate: among 69 intact-image answer changes, only 3 recur under text-only input and 7 under shuffled images, so the positive control gains arise mostly on different examples. Geometry remains sparse (3,206 eligible pairs), and reciprocal consistency refinement is a smaller conflict-selection refinement.

\paragraph{Model and compute controls.}
Qwen3-I is stronger on Full, but Qwen3-T receives the larger complete-pool gain (+4.95 versus +3.03). Closed-set scoring does not isolate free-form deliberation, so this shows only that a reasoning checkpoint can benefit from explicit evidence. On 3,600 balanced rows, seven Full prompts reach 70.25 by score averaging and 70.28 by voting at $7.00\times$ cost, versus SEER's 74.64 at $7.12\times$. A post-hoc frozen efficiency diagnostic invokes full SEER only below a model-specific 25th-percentile Full-score gap selected on unlabeled development data. It covers 25.7\% of rows, obtains a pooled +3.22 [1.89,4.61], and retains 81.7\% of the full +3.94 gain at $2.57\times$ rather than $7.12\times$ cost. We report this as an accuracy--cost diagnostic, not a second primary method.

\paragraph{Robustness, grounding, and scope.}
The method best fits named pairs with relations visible from local appearance or 2D layout. Hash improves five relations and regresses on four; the worst pooled gains are behind $-4.5$, contains/on $-4.0$, and right-of $-2.5$. Removing left/right, inside/contains, role-swapped, or target-\emph{on} rows still yields subject-first gains of +3.57, +2.64, +4.13, and +4.75. On the complete frozen set, 535 unique-name and 365 repeated-name images retain pooled gains of +2.24 [0.37,4.11] and +3.42 [0.41,6.44]. In the targeted audit, unanimously identifiable and non-unique images retain +8.70 [2.17,17.39] and +6.38 [$-4.26$,17.02]; the 78-item strict-consensus subset retains +5.77 [$-1.28$,13.46]. Gains are therefore not confined to ambiguous pairs, although SEER remains an evidence interface rather than a certificate of exact grounding.

\section{Limitations}

\method{} is bounded by grounding quality and relations visible from local appearance or 2D layout. Relation hiding does not remove order bias, although counterbalanced gains remain positive. Reciprocal refinement requires inverse-supported relation choice and binary transfer is inconsistent. Full SEER costs $7.12\times$ one Full score; its cheaper gate is post-hoc. Image-disjoint freezing cannot exclude pretraining exposure, and option-free evidence is tested only with Qwen3-I.

\section{Conclusion}

\method{} turns relation-hidden self-grounding into role-explicit evidence without training a new model or using an external detector. Across frozen and cross-model evaluations, matched controls identify local refocus and semantic S/O binding as the principal source of improvement, while reciprocal consistency provides a smaller refinement when exact inverses are available. These results show that explicit evidence interfaces can complement stronger reasoning checkpoints, although robust grounding and open-ended spatial semantics remain unresolved.

\bibliography{spatial_evidence}

@article{krishna2017visualgenome,
  title={Visual Genome: Connecting Language and Vision Using Crowdsourced Dense Image Annotations},
  author={Krishna, Ranjay and Zhu, Yuke and Groth, Oliver and Johnson, Justin and Hata, Kenji and Kravitz, Joshua and Chen, Stephanie and Kalantidis, Yannis and Li, Li-Jia and Shamma, David A. and Bernstein, Michael S. and Fei-Fei, Li},
  journal={International Journal of Computer Vision},
  volume={123},
  number={1},
  pages={32--73},
  year={2017}
}

@inproceedings{hudson2019gqa,
  title={GQA: A New Dataset for Real-World Visual Reasoning and Compositional Question Answering},
  author={Hudson, Drew A. and Manning, Christopher D.},
  booktitle={Proceedings of the IEEE/CVF Conference on Computer Vision and Pattern Recognition},
  pages={6700--6709},
  year={2019}
}

@inproceedings{lin2014coco,
  title={Microsoft COCO: Common Objects in Context},
  author={Lin, Tsung-Yi and Maire, Michael and Belongie, Serge and Hays, James and Perona, Pietro and Ramanan, Deva and Dollar, Piotr and Zitnick, C. Lawrence},
  booktitle={European Conference on Computer Vision},
  pages={740--755},
  year={2014}
}

@article{liu2023vsr,
  title={Visual Spatial Reasoning},
  author={Liu, Fangyu and Emerson, Guy and Collier, Nigel},
  journal={Transactions of the Association for Computational Linguistics},
  volume={11},
  pages={635--651},
  year={2023},
  doi={10.1162/tacl_a_00566}
}

@inproceedings{du2024embspatial,
  title={EmbSpatial-Bench: Benchmarking Spatial Understanding for Embodied Tasks with Large Vision-Language Models},
  author={Du, Mengfei and Wu, Binhao and Li, Zejun and Huang, Xuanjing and Wei, Zhongyu},
  booktitle={Proceedings of the 62nd Annual Meeting of the Association for Computational Linguistics (Volume 2: Short Papers)},
  year={2024},
  publisher={Association for Computational Linguistics},
  url={https://aclanthology.org/2024.acl-short.33/}
}

@inproceedings{kamath2023whatsup,
  title={What's ``up'' with Vision-Language Models? Investigating Their Struggle with Spatial Reasoning},
  author={Kamath, Amita and Hessel, Jack and Chang, Kai-Wei},
  booktitle={Proceedings of the 2023 Conference on Empirical Methods in Natural Language Processing},
  pages={9161--9175},
  year={2023}
}

@inproceedings{yuksekgonul2023bags,
  title={When and Why Vision-Language Models Behave like Bags-of-Words, and What to Do about It?},
  author={Yuksekgonul, Mert and Bianchi, Federico and Kalluri, Pratyusha and Jurafsky, Dan and Zou, James},
  booktitle={International Conference on Learning Representations},
  year={2023}
}

@inproceedings{hsieh2023sugarcrepe,
  title={SugarCrepe: Fixing Hackable Benchmarks for Vision-Language Compositionality},
  author={Hsieh, Cheng-Yu and Zhang, Jieyu and Ma, Zixian and Kembhavi, Aniruddha and Krishna, Ranjay},
  booktitle={Advances in Neural Information Processing Systems},
  year={2023}
}

@inproceedings{guan2024hallusionbench,
  title={HallusionBench: An Advanced Diagnostic Suite for Entangled Language Hallucination and Visual Illusion in Large Vision-Language Models},
  author={Guan, Tianrui and Liu, Fuxiao and Wu, Xiyang and Xian, Ruiqi and Li, Zongxia and Liu, Xiaoyu and Wang, Xijun and Chen, Lichang and Huang, Furong and Yacoob, Yaser and Manocha, Dinesh and Zhou, Tianyi},
  booktitle={Proceedings of the IEEE/CVF Conference on Computer Vision and Pattern Recognition},
  pages={14375--14385},
  year={2024}
}

@misc{yang2023som,
  title={Set-of-Mark Prompting Unleashes Extraordinary Visual Grounding in GPT-4V},
  author={Yang, Jianwei and Zhang, Hao and Li, Feng and Zou, Xueyan and Li, Chunyuan and Gao, Jianfeng},
  year={2023},
  eprint={2310.11441},
  archivePrefix={arXiv},
  primaryClass={cs.CV}
}

@article{frisoni2026graphofmark,
  title={Graph-of-Mark: Promote Spatial Reasoning in Multimodal Language Models with Graph-Based Visual Prompting},
  author={Frisoni, Giacomo and Molfetta, Lorenzo and Buzzoni, Mattia and Moro, Gianluca},
  journal={Proceedings of the AAAI Conference on Artificial Intelligence},
  year={2026},
  volume={40},
  number={36},
  pages={30726--30734},
  doi={10.1609/aaai.v40i36.40329},
  url={https://ojs.aaai.org/index.php/AAAI/article/view/40329}
}

@article{xu2026vpbench,
  title={VP-Bench: A Comprehensive Benchmark for Visual Prompting in Multimodal Large Language Models},
  author={Xu, Mingjie and Chen, Jinpeng and Zhao, Yuzhi and Li, Jason Chun Lok and Qiu, Yue and Du, Zekang and Wu, Mengyang and Zhang, Pingping and Li, Kun and Yang, Hongzheng and Ma, Wenao and Wei, Jiaheng and Li, Qinbin and Liu, Kangcheng and Lei, Wenqiang},
  journal={Proceedings of the AAAI Conference on Artificial Intelligence},
  year={2026},
  volume={40},
  number={13},
  pages={11332--11341},
  doi={10.1609/aaai.v40i13.38114},
  url={https://ojs.aaai.org/index.php/AAAI/article/view/38114}
}

@inproceedings{chou2026testtime,
  title={Test-Time Consistency in Vision Language Models},
  author={Chou, Shih-Han and Chandhok, Shivam and Little, James J. and Sigal, Leonid},
  booktitle={Proceedings of the IEEE/CVF Winter Conference on Applications of Computer Vision},
  year={2026},
  pages={7789--7798},
  url={https://openaccess.thecvf.com/content/WACV2026/html/Chou_Test-Time_Consistency_in_Vision_Language_Models_WACV_2026_paper.html}
}

@inproceedings{cheng2024spatialrgpt,
  title={SpatialRGPT: Grounded Spatial Reasoning in Vision Language Models},
  author={Cheng, An-Chieh and Yin, Hongxu and Fu, Yang and Guo, Qiushan and Yang, Ruihan and Kautz, Jan and Wang, Xiaolong and Liu, Sifei},
  booktitle={Advances in Neural Information Processing Systems},
  year={2024}
}

@misc{bai2025qwen25vl,
  title={Qwen2.5-VL Technical Report},
  author={Bai, Shuai and Chen, Keqin and Liu, Xuejing and Wang, Jialin and Ge, Wenbin and Song, Sibo and Dang, Kai and Wang, Peng and Wang, Shijie and Tang, Jun and Zhong, Humen and Zhu, Yuanzhi and Yang, Mingkun and Li, Zhaohai and Wan, Jianqiang and Wang, Pengfei and Ding, Wei and Fu, Zheren and Xu, Yiheng and Ye, Jiabo and Zhang, Xi and Xie, Tianbao and Cheng, Zesen and Zhang, Hang and Yang, Zhibo and Xu, Haiyang and Lin, Junyang},
  year={2025},
  eprint={2502.13923},
  archivePrefix={arXiv},
  primaryClass={cs.CV}
}

@misc{bai2025qwen3vl,
  title={Qwen3-VL Technical Report},
  author={Bai, Shuai and Chen, Keqin and others},
  year={2025},
  eprint={2511.21631},
  archivePrefix={arXiv},
  primaryClass={cs.CV}
}

@inproceedings{yang2025r1onevision,
  title={R1-Onevision: Advancing Generalized Multimodal Reasoning through Cross-Modal Formalization},
  author={Yang, Yi and He, Xiaoxuan and Pan, Hongkun and Jiang, Xiyan and Deng, Yan and Yang, Xingtao and Lu, Haoyu and Yin, Dacheng and Rao, Fengyun and Zhu, Minfeng and Zhang, Bo and Chen, Wei},
  booktitle={Proceedings of the IEEE/CVF International Conference on Computer Vision},
  year={2025},
  pages={2376--2385}
}

@inproceedings{liu2024llava15,
  title={Improved Baselines with Visual Instruction Tuning},
  author={Liu, Haotian and Li, Chunyuan and Li, Yuheng and Lee, Yong Jae},
  booktitle={Proceedings of the IEEE/CVF Conference on Computer Vision and Pattern Recognition},
  pages={26296--26306},
  year={2024}
}

@misc{wang2025internvl35,
  title={InternVL3.5: Advancing Open-Source Multimodal Models in Versatility, Reasoning, and Efficiency},
  author={Wang, Weiyun and others},
  year={2025},
  eprint={2510.17707},
  archivePrefix={arXiv},
  primaryClass={cs.CV}
}

@inproceedings{lu2026reallocating,
  title={Reallocating Attention Across Layers to Reduce Multimodal Hallucination},
  author={Lu, Haolang and Chu, Bolun and Fu, WeiYe and Nan, Guoshun and Liu, Junning and Pan, Minghui and Li, Qiankun and Yu, Yi and Wang, Hua and Wang, Kun},
  booktitle={Proceedings of the IEEE/CVF Conference on Computer Vision and Pattern Recognition},
  year={2026},
  pages={4157--4167}
}

@misc{yu2025sibench,
  title={How Far are VLMs from Visual Spatial Intelligence? A Benchmark-Driven Perspective},
  author={Yu, Songsong and Chen, Yuxin and Ju, Hao and Jia, Lianjie and Zhang, Fuxi and Huang, Shaofei and Wu, Yuhan and Cui, Rundi and Ran, Binghao and Zhang, Zaibin and Zheng, Zhedong and Zhang, Zhipeng and Wang, Yifan and Song, Lin and Wang, Lijun and Li, Yanwei and Shan, Ying and Lu, Huchuan},
  year={2025},
  eprint={2509.18905},
  archivePrefix={arXiv},
  primaryClass={cs.CV}
}

@inproceedings{antol2015vqa,
  title={VQA: Visual Question Answering},
  author={Antol, Stanislaw and Agrawal, Aishwarya and Lu, Jiasen and Mitchell, Margaret and Batra, Dhruv and Zitnick, C. Lawrence and Parikh, Devi},
  booktitle={Proceedings of the IEEE International Conference on Computer Vision},
  pages={2425--2433},
  year={2015}
}

@inproceedings{goyal2017vqa,
  title={Making the V in VQA Matter: Elevating the Role of Image Understanding in Visual Question Answering},
  author={Goyal, Yash and Khot, Tejas and Summers-Stay, Douglas and Batra, Dhruv and Parikh, Devi},
  booktitle={Proceedings of the IEEE Conference on Computer Vision and Pattern Recognition},
  pages={6904--6913},
  year={2017}
}

@inproceedings{johnson2017clevr,
  title={CLEVR: A Diagnostic Dataset for Compositional Language and Elementary Visual Reasoning},
  author={Johnson, Justin and Hariharan, Bharath and van der Maaten, Laurens and Fei-Fei, Li and Zitnick, C. Lawrence and Girshick, Ross},
  booktitle={Proceedings of the IEEE Conference on Computer Vision and Pattern Recognition},
  pages={2901--2910},
  year={2017}
}

@inproceedings{suhr2019nlvr2,
  title={A Corpus for Reasoning about Natural Language Grounded in Photographs},
  author={Suhr, Alane and Zhou, Stephanie and Zhang, Ally and Zhang, Iris and Bai, Huajun and Artzi, Yoav},
  booktitle={Proceedings of the 57th Annual Meeting of the Association for Computational Linguistics},
  pages={6418--6428},
  year={2019}
}

@inproceedings{thrush2022winoground,
  title={Winoground: Probing Vision and Language Models for Visio-Linguistic Compositionality},
  author={Thrush, Tristan and Jiang, Ryan and Bartolo, Max and Singh, Amanpreet and Williams, Adina and Kiela, Douwe and Ross, Candace},
  booktitle={Proceedings of the IEEE/CVF Conference on Computer Vision and Pattern Recognition},
  pages={5238--5248},
  year={2022}
}

@inproceedings{parcalabescu2022valse,
  title={VALSE: A Task-Independent Benchmark for Vision and Language Models Centered on Linguistic Phenomena},
  author={Parcalabescu, Letitia and Cafagna, Michele and Muradjan, Lilitta and Frank, Anette and Calixto, Iacer and Gatt, Albert},
  booktitle={Proceedings of the 60th Annual Meeting of the Association for Computational Linguistics},
  pages={8253--8280},
  year={2022}
}

@misc{rajabi2024gsrbench,
  title={GSR-Bench: A Benchmark for Grounded Spatial Reasoning Evaluation via Multimodal LLMs},
  author={Rajabi, Navid and Kosecka, Jana},
  year={2024},
  eprint={2406.13246},
  archivePrefix={arXiv},
  primaryClass={cs.CV}
}

@inproceedings{chen2024spatialvlm,
  title={SpatialVLM: Endowing Vision-Language Models with Spatial Reasoning Capabilities},
  author={Chen, Boyuan and Xu, Zhuo and Kirmani, Sean and Ichter, Brian and Driess, Danny and Florence, Pete and Sadigh, Dorsa and Guibas, Leonidas and Xia, Fei},
  booktitle={Proceedings of the IEEE/CVF Conference on Computer Vision and Pattern Recognition},
  pages={14455--14465},
  year={2024}
}

@misc{jia2025omnispatial,
  title={OmniSpatial: Towards Comprehensive Spatial Reasoning Benchmark for Vision Language Models},
  author={Jia, Mengdi and Qi, Zekun and Zhang, Shaochen and Zhang, Wenyao and Yu, Xinqiang and He, Jiawei and Wang, He and Yi, Li},
  year={2025},
  eprint={2506.03135},
  archivePrefix={arXiv},
  primaryClass={cs.CV}
}

@inproceedings{radford2021clip,
  title={Learning Transferable Visual Models From Natural Language Supervision},
  author={Radford, Alec and Kim, Jong Wook and Hallacy, Chris and Ramesh, Aditya and Goh, Gabriel and Agarwal, Sandhini and Sastry, Girish and Askell, Amanda and Mishkin, Pamela and Clark, Jack and Krueger, Gretchen and Sutskever, Ilya},
  booktitle={Proceedings of the 38th International Conference on Machine Learning},
  pages={8748--8763},
  year={2021}
}

@inproceedings{li2022blip,
  title={BLIP: Bootstrapping Language-Image Pre-training for Unified Vision-Language Understanding and Generation},
  author={Li, Junnan and Li, Dongxu and Xiong, Caiming and Hoi, Steven},
  booktitle={Proceedings of the 39th International Conference on Machine Learning},
  pages={12888--12900},
  year={2022}
}

@inproceedings{li2023blip2,
  title={BLIP-2: Bootstrapping Language-Image Pre-training with Frozen Image Encoders and Large Language Models},
  author={Li, Junnan and Li, Dongxu and Savarese, Silvio and Hoi, Steven},
  booktitle={Proceedings of the 40th International Conference on Machine Learning},
  year={2023}
}

@inproceedings{alayrac2022flamingo,
  title={Flamingo: A Visual Language Model for Few-Shot Learning},
  author={Alayrac, Jean-Baptiste and Donahue, Jeff and Luc, Pauline and Miech, Antoine and Barr, Iain and Hasson, Yana and Lenc, Karel and Mensch, Arthur and Millican, Katie and Reynolds, Malcolm and Ring, Roman and Rutherford, Eliza and Cabi, Serkan and Han, Tengda and Gong, Zhitao and Samangooei, Sina and Monteiro, Marianne and Menick, Jacob and Borgeaud, Sebastian and Brock, Andy and Nematzadeh, Aida and Sharifzadeh, Sahand and Binkowski, Mikolaj and Barreira, Ricardo and Vinyals, Oriol and Zisserman, Andrew and Simonyan, Karen},
  booktitle={Advances in Neural Information Processing Systems},
  year={2022}
}

@inproceedings{dai2023instructblip,
  title={InstructBLIP: Towards General-Purpose Vision-Language Models with Instruction Tuning},
  author={Dai, Wenliang and Li, Junnan and Li, Dongxu and Tiong, Anthony Meng Huat and Zhao, Junqi and Wang, Weisheng and Li, Boyang and Fung, Pascale and Hoi, Steven},
  booktitle={Advances in Neural Information Processing Systems},
  year={2023}
}

@inproceedings{peng2024kosmos2,
  title={Kosmos-2: Grounding Multimodal Large Language Models to the World},
  author={Peng, Zhiliang and Wang, Wenhui and Dong, Li and Hao, Yaru and Huang, Shaohan and Ma, Shuming and Wei, Furu},
  booktitle={International Conference on Learning Representations},
  year={2024}
}

@inproceedings{li2023pope,
  title={Evaluating Object Hallucination in Large Vision-Language Models},
  author={Li, Yifan and Du, Yifan and Zhou, Kun and Wang, Jinpeng and Zhao, Wayne Xin and Wen, Ji-Rong},
  booktitle={Proceedings of the 2023 Conference on Empirical Methods in Natural Language Processing},
  pages={292--305},
  year={2023}
}

@inproceedings{liu2024groundingdino,
  title={Grounding DINO: Marrying DINO with Grounded Pre-Training for Open-Set Object Detection},
  author={Liu, Shilong and Zeng, Zhaoyang and Ren, Tianhe and Li, Feng and Zhang, Hao and Yang, Jie and Jiang, Qing and Li, Chunyuan and Yang, Jianwei and Su, Hang and Zhu, Jun and Zhang, Lei},
  booktitle={European Conference on Computer Vision},
  year={2024}
}

@inproceedings{minderer2022owlvit,
  title={Simple Open-Vocabulary Object Detection with Vision Transformers},
  author={Minderer, Matthias and Gritsenko, Alexey and Stone, Austin and Neumann, Maxim and Weissenborn, Dirk and Dosovitskiy, Alexey and Mahendran, Aravindh and Arnab, Anurag and Dehghani, Mostafa and Shen, Zhuoran and Wang, Xiao and Zhai, Xiaohua and Kipf, Thomas and Houlsby, Neil},
  booktitle={European Conference on Computer Vision},
  pages={728--755},
  year={2022}
}

@inproceedings{kirillov2023sam,
  title={Segment Anything},
  author={Kirillov, Alexander and Mintun, Eric and Ravi, Nikhila and Mao, Hanzi and Rolland, Chloe and Gustafson, Laura and Xiao, Tete and Whitehead, Spencer and Berg, Alexander C. and Lo, Wan-Yen and Dollar, Piotr and Girshick, Ross},
  booktitle={Proceedings of the IEEE/CVF International Conference on Computer Vision},
  pages={4015--4026},
  year={2023}
}

@article{chen2023shikra,
  title={Shikra: Unleashing Multimodal LLM's Referential Dialogue Magic},
  author={Chen, Keqin and Zhang, Zhao and Zeng, Weili and Zhang, Richong and Zhu, Feng and Zhao, Rui},
  journal={arXiv preprint arXiv:2306.15195},
  year={2023}
}

@inproceedings{you2024ferret,
  title={Ferret: Refer and Ground Anything Anywhere at Any Granularity},
  author={You, Haoxuan and Zhang, Haotian and Gan, Zhe and Du, Xianzhi and Zhang, Bowen and Wang, Zirui and Cao, Liangliang and Chang, Shih-Fu and Yang, Yinfei},
  booktitle={International Conference on Learning Representations},
  year={2024}
}

@inproceedings{wei2022cot,
  title={Chain-of-Thought Prompting Elicits Reasoning in Large Language Models},
  author={Wei, Jason and Wang, Xuezhi and Schuurmans, Dale and Bosma, Maarten and Xia, Fei and Chi, Ed and Le, Quoc V. and Zhou, Denny},
  booktitle={Advances in Neural Information Processing Systems},
  year={2022}
}

@inproceedings{chou2025mmr3,
  title = {{MM}-R$^3$: On (In-)Consistency of Vision-Language Models ({VLM}s)},
  author = {Chou, Shih-Han and Chandhok, Shivam and Little, Jim and Sigal, Leonid},
  booktitle = {Findings of the Association for Computational Linguistics: ACL 2025},
  pages = {4762--4788},
  year = {2025},
  publisher = {Association for Computational Linguistics},
  doi = {10.18653/v1/2025.findings-acl.246},
  url = {https://aclanthology.org/2025.findings-acl.246/}
}

@inproceedings{yang2019spatialsense,
  title={SpatialSense: An Adversarially Crowdsourced Benchmark for Spatial Relation Recognition},
  author={Yang, Kaiyu and Russakovsky, Olga and Deng, Jia},
  booktitle={Proceedings of the IEEE/CVF International Conference on Computer Vision},
  year={2019},
  doi={10.1109/ICCV.2019.00214}
}

\clearpage
\section*{Supplementary Material}
\addcontentsline{toc}{section}{Supplementary Material}
\raggedbottom

\section{Target-Complete Evaluation Protocol}

\subsection{Construction and Shortcut Audit}

The target-complete development suites are derived from GQA \citep{hudson2019gqa} and Visual Genome (VG) \citep{krishna2017visualgenome} scene graphs. We canonicalize relation names, require valid images and subject/object annotations, and deduplicate tuples. Four balanced option families cover left/right, above/below/on, front/behind, and inside/contains; every displayed relation therefore also occurs as a target.

We exclude proximity because the source graphs contain many ``near'' annotations but almost no reliable ``far'' targets. For containment, we partition the available ``inside'' examples into two non-overlapping subsets and role-swap one subset to form ``contains.'' Candidate order is shuffled with seed 2027.

\begin{table}[!ht]
\centering
\small
\setlength{\tabcolsep}{2pt}
\begin{tabular*}{\columnwidth}{@{\extracolsep{\fill}}lrrr@{}}
\toprule
Relation family & GQA & VG & Option prior \\
\midrule
Left / right & 1,500 each & 351 each & 50.00 \\
Above / below / on & 1,190 each & 1,500 each & 33.33 \\
Front / behind & 1,500 each & 1,500 each & 50.00 \\
Inside / contains & 750 each & 750 each & 50.00 \\
\midrule
Total & 11,070 & 9,702 & -- \\
\bottomrule
\end{tabular*}
\caption{Target-complete relation counts. Option prior is the majority-target accuracy within that answer set; all displayed relations are balanced targets.}
\label{tab:sup_dataset_counts}
\end{table}

The cross-model sets sample 100 examples for each of nine labels (900 per dataset) from these pools before model evaluation. Full-scale evaluation measures the available relation distribution; the balanced sets give each relation equal weight across evaluated VLMs.

\subsection{Frozen Same-Source, Image-Disjoint Test}

To separate confirmation from protocol development, we independently derive GQA-Train900 from GQA training scene graphs. We first remove every image ID appearing in either GQA or VG development pool, then select 100 examples for each of the same nine target relations while allowing at most one tuple per image. The resulting test has 900 examples, 900 images, and no development-image overlap; 535 examples contain a unique named subject and object instance in the source scene graph, while 365 contain at least one repeated name. Selection uses only sample ID, image ID, and target relation---never model outputs, boxes, option scores, or correctness.

Before scoring, we fixed the exact 900 sample IDs, source-file checksums, selection seed 20270723, relation counts, and output SHA256 (prefix \texttt{cc1319b3d942}). The $\tau=0.5$ forward rule and reciprocal unique-consistency rule had both been frozen earlier, so this set supplies an untouched test of the complete nine-relation interface rather than another source for method selection.

\begin{table}[!ht]
\centering
\small
\setlength{\tabcolsep}{1.2pt}
\begin{tabular*}{\columnwidth}{@{\extracolsep{\fill}}lrrrrrr@{}}
\toprule
Model & Full & Marker & Fwd. & \method{} & Gain & $\mathrm{CI}_{img}$ \\
\midrule
Qwen3-I & 75.33 & 76.11 & 77.44 & \textbf{78.44} & +3.11 & [1.00,5.22] \\
Qwen3-T & 70.56 & 75.22 & 73.89 & \textbf{75.33} & +4.78 & [2.56,7.11] \\
\midrule
Both & 72.94 & 75.67 & 75.67 & \textbf{76.89} & +3.94 & [2.17,5.72] \\
\bottomrule
\end{tabular*}
\caption{Frozen GQA-Train900 results. Fwd. is the pre-refinement output; final SEER applies the previously frozen reciprocal rule.}
\label{tab:sup_frozen_gqa_train}
\end{table}

Against the self-grounded Marker state, final SEER gains +2.33 points for Qwen3-I (36 fixes/15 breaks, McNemar $p=.0046$, image-cluster 95\% CI [+0.78,+3.89]) and +0.11 for Qwen3-T (26/25, $p=1.0$, [$-1.44$,+1.67]). Pooling the two model outcomes gives +1.22 [+0.06,+2.33]. Thus the frozen aggregate supports a decision benefit beyond evidence construction, while the per-model intervals show that its magnitude is heterogeneous.

\begin{table}[t]
\centering
\small
\setlength{\tabcolsep}{1.15pt}
\begin{tabular*}{\columnwidth}{@{\extracolsep{\fill}}lrrrrrr@{}}
\toprule
& \multicolumn{3}{c}{Qwen3-I} & \multicolumn{3}{c}{Qwen3-T} \\
\cmidrule(lr){2-4}\cmidrule(lr){5-7}
Relation & Full & SEER & Gain & Full & SEER & Gain \\
\midrule
above & 61 & 64 & +3 & 56 & 60 & +4 \\
behind & 86 & 81 & $-5$ & 85 & 81 & $-4$ \\
below & 64 & 70 & +6 & 65 & 68 & +3 \\
contains & 92 & 96 & +4 & 91 & 92 & +1 \\
in front of & 74 & 85 & +11 & 68 & 71 & +3 \\
inside & 82 & 88 & +6 & 64 & 87 & +23 \\
left of & 58 & 72 & +14 & 60 & 80 & +20 \\
on & 80 & 75 & $-5$ & 77 & 77 & 0 \\
right of & 81 & 75 & $-6$ & 69 & 62 & $-7$ \\
\bottomrule
\end{tabular*}
\caption{Per-relation accuracy on the frozen nine-relation test. Each model has exactly 100 examples per row.}
\label{tab:sup_frozen_gqa_train_relation}
\end{table}

Because the largest frozen loss is on right-of, we recompute the Forward outputs after disabling Geometry and routing only between Full and Marker by the same score-gap rule. Table~\ref{tab:sup_frozen_no_geometry} shows that Geometry is not the main cause of the pre-refinement asymmetry. Reciprocal consistency refinement raises Qwen3-I right-of from 71 to 75 but changes Qwen3-T from 63 to 62, so it reduces the pooled deficit without eliminating grounding-order and interface bias.

\begin{table}[t]
\centering
\small
\setlength{\tabcolsep}{2pt}
\begin{tabular*}{\columnwidth}{@{\extracolsep{\fill}}llrrrr@{}}
\toprule
Model & Relation & Full & Marker & Visual only & Fwd. \\
\midrule
Qwen3-I & left of & 58 & 73 & 72 & 72 \\
Qwen3-I & right of & 81 & 67 & 71 & 71 \\
Qwen3-T & left of & 60 & 84 & 72 & 73 \\
Qwen3-T & right of & 69 & 65 & 63 & 63 \\
\bottomrule
\end{tabular*}
\caption{Frozen directional diagnostic with Geometry disabled. ``Visual only'' compares Full and Marker option-score gaps; all rows contain 100 examples.}
\label{tab:sup_frozen_no_geometry}
\end{table}

Forward improves Qwen3-I on both source-scene multiplicity strata: +1.68 on 535 unique-instance examples and +2.74 on 365 examples with a repeated subject or object name. Qwen3-T gains +2.80 and +4.11 on the same strata. Relative to source boxes, Qwen3-I/Qwen3-T grounding succeeds on 892/886 examples and has mean subject/object IoU 0.421/0.432 and 0.380/0.386. Positive Forward gains persist even when the smaller S/O IoU is zero (+1.40/+2.37), confirming that source-box overlap is an imperfect proxy and should not be read as a human pair-correctness estimate.

\subsection{Grounder and Evaluated VLM}

The evaluated VLM $M$ and VLM grounder $G$ may differ. Qwen3-I/T use $G=M$; Qwen2.5-I and R1-OV use Qwen3-I as $G$. Direct R1-OV grounding returned only 1/30 valid boxes, so we do not call its transfer setting self-grounded.

$G$ receives the image, subject/object phrases, and the generic relation question only to resolve the relevant instances; answer options, candidate relation labels, and any request to produce an answer are withheld. It returns normalized JSON boxes. Boxes are clipped and rejected for non-positive area, side below 0.005, area below $10^{-5}$, or extreme aspect ratio. Invalid grounding falls back to Full.

\subsection{State Scoring and Evidence Decision}

Full and Marker use the same frozen $M$, candidates, option order, and length-normalized continuation log-probability; the option-score gap is the difference between the top two scores. Marker crops the padded box union, draws red S and blue O boxes, and states those identities in the prompt. For a left/right-only candidate family, Geometry predicts left when $d_x=x_s-x_o<0$ and right otherwise, with layout score $c=|d_x|$, where $x_s$ and $x_o$ are normalized box-center coordinates. For an inside/contains-only family, let $u_s=|b_s\cap b_o|/|b_s|$ and $u_o=|b_s\cap b_o|/|b_o|$; Geometry predicts inside when $u_s>u_o$ and contains otherwise, with $c=|u_s-u_o|$. It abstains on above/below/on and front/behind. At fixed $\tau=0.5$, eligible Geometry takes priority; otherwise the larger-gap visual state defines Forward. No decision component is fit to labels.

Final SEER follows the main-paper reciprocal rule only for inverse-supported relation-choice protocols. It exchanges entity roles and visible S/O identities while preserving option strings and order. On an eligible Full/Marker conflict, the forward source changes only when exactly one state predicts the inverse of its own forward relation; a missing inverse candidate counts as inconsistent. Both- or neither-consistent cases retain Forward. SpatialSense below bounds this relation-MC rule outside relation classification.

Option-score gaps provide an ordering cue but are not calibrated state-correctness probabilities. Across the four complete-pool Qwen3 runs, their state-selection AUC is only .565--.602; Qwen3-I Marker gaps exceed Full by roughly three points and win 76--77\% of raw comparisons, whereas Qwen3-T scales are nearly matched. The highest equal-frequency gap bin is nevertheless 6.92--21.25 points more accurate than the lowest. Label-free z-score, median/IQR, and empirical-percentile normalizations change frozen GQA and EmbSpatial accuracy by at most 0.41 points and do not dominate raw gaps. We therefore retain the transparent raw comparison in Forward and use reciprocal structure for the final refinement.

\begin{table*}[t]
\centering
\small
\setlength{\tabcolsep}{3pt}
\begin{tabular*}{\textwidth}{@{\extracolsep{\fill}}llrrrrr@{}}
\toprule
Protocol & Model & N & Full & Marker & Fwd. & \method{} \\
\midrule
Frozen GQA-Train900 & Qwen3-I & 900 & 75.33 & 76.11 & 77.44 & \textbf{78.44} \\
 & Qwen3-T & 900 & 70.56 & 75.22 & 73.89 & \textbf{75.33} \\
 & Pooled & 1,800 & 72.94 & 75.67 & 75.67 & \textbf{76.89} \\
\midrule
GQA/VG development & Qwen3-I & 20,772 & 72.62 & 73.90 & 74.43 & \textbf{75.65} \\
 & Qwen3-T & 20,772 & 68.59 & 71.67 & 72.05 & \textbf{73.54} \\
 & Pooled & 41,544 & 70.60 & 72.79 & 73.24 & \textbf{74.60} \\
\midrule
Frozen EmbSpatial & Qwen3-I & 2,434 & 81.88 & 85.33 & 86.24 & \textbf{86.24} \\
 & Qwen3-T & 2,434 & 78.64 & 82.95 & 83.48 & \textbf{83.73} \\
 & InternVL3.5 & 2,434 & 65.74 & \textbf{78.06} & 77.28 & 77.53 \\
 & Pooled & 7,302 & 75.42 & 82.11 & 82.33 & \textbf{82.50} \\
\bottomrule
\end{tabular*}
\caption{Evidence construction and decision decomposition. Fwd. denotes Geometry/margin composition before reciprocal consistency refinement; final SEER is the primary relation-MC output.}
\label{tab:sup_decision_increment}
\end{table*}

Unconditional Full/Marker fusion is similarly unhelpful on 41,544 development evaluations: always-Marker gives 72.79, raw-gap selection 72.73, mean-probability/log-score fusion 72.80/72.83, and rank fusion 71.91, versus 73.24 for Forward and 74.60 for final SEER. The Full/Marker oracle is 80.64, so complementarity exists but is not recovered by global score averaging.

An exact No-Geometry ablation further isolates the sparse box-derived route before reciprocal refinement. On frozen Qwen3-I/T, No-Geometry Forward versus Forward is 77.44/77.44 and 73.44/73.89, giving a pooled Geometry increment of +0.22 [$-0.17$,0.67]. On the four complete development rows, the increments are +0.19 GQA and +0.10 VG for Qwen3-I, and +1.05/+0.68 for Qwen3-T; pooled over 41,544 evaluations, Forward improves No-Geometry Forward by +0.51 [0.40,0.63], with 309 fixes, 96 breaks, and 3,206 Geometry routes. EmbSpatial has no eligible Geometry route under the frozen protocol. Thus Geometry is a modest positive development component, not the main source of improvement.

\subsection{Prompts, Rendering, and Scoring}

Table \ref{tab:sup_prompts} gives the target-complete prompt skeletons. Grounding retains the entity names and generic question only to resolve instances; options and relation labels are removed. Full and Marker use the same question, options, and order, with Marker adding only visible S/O role cues and the corresponding instruction.

\begin{table*}[t]
\centering
\small
\setlength{\tabcolsep}{2pt}
\begin{tabular}{p{0.15\textwidth}p{0.80\textwidth}}
\toprule
Stage & Prompt skeleton \\
\midrule
Grounding & Locate queried S and O in the full image without answering. Return only valid JSON with normalized $[0,1]$ \texttt{subject\_bbox} and \texttt{object\_bbox}; choose the pair relevant to $q$. Inputs are $s$, $o$, and $q$, with options removed. \\
Full scoring & Present $q$ and the fixed ordered options; request exactly one option letter $\mathcal{L}$. \\
Marker scoring & State that red S marks $s$ and blue O marks $o$; present the same $q$ and options; specify relation direction S-to-O, ignore unmarked duplicates, and request one letter $\mathcal{L}$. \\
Reciprocal scoring & Exchange $s$ and $o$ in $q$ while preserving option strings and order; for Marker, also exchange visible S/O assignments. \\
Open output & Remove options and request one spatial-relation phrase without explanation, coordinates, or an option letter. \\
\bottomrule
\end{tabular}
\caption{Prompt skeletons. $q$, $s$, $o$, and $\mathcal{L}$ denote the sample question, subject, object, and ordered option labels. Quotation is compacted for presentation while preserving all task-relevant content.}
\label{tab:sup_prompts}
\end{table*}

Boxes are clipped to the image and validated before rendering. For valid boxes, Marker crops the union of the pair and expands it on each side by $0.7$ times the union width or height, subject to image boundaries. The subject is rendered with a red rectangle and an S label; the object uses blue and O. If grounding is invalid, Marker and Geometry are unavailable and the decision is exactly the Full prediction. Across all 41,544 model--example evaluations in the complete target-complete pools, each of the 440 invalid-grounding cases follows this Full route. Marker scores for this subset are unused candidates; the routed output is identical to Full.

For each visual state, the frozen VLM evaluates every option continuation and records its mean token negative log-likelihood. Prediction selects the minimum-NLL option, equivalently the maximum length-normalized log-probability, and the state margin is the gap between the best and second-best scores. No option is generated and reparsed for the closed-set tables. Forward Full/Marker pairs and their reciprocal queries retain identical option strings and option order; only the queried entity roles, and the visible S/O identities for Marker, are exchanged.

\begin{table}[!ht]
\centering
\small
\begin{tabular*}{\columnwidth}{@{\extracolsep{\fill}}lrrr@{}}
\toprule
Protocol & Ground calls & Score calls & Learned params. \\
\midrule
Full & 0 & 1 & 0 \\
Marker & 1 & 1 & 0 \\
Forward & 1 & 2 & 0 \\
\method{} & 1 & $2+2\mathbb{1}[\mathrm{recip.}]$ & 0 \\
\bottomrule
\end{tabular*}
\caption{Per-example call accounting when Full has not been precomputed. Reciprocal scores are requested only for eligible visual-state disagreements (10.27\% in the complete-pool study).}
\label{tab:sup_cost}
\end{table}

\begin{figure}[t]
\centering
\includegraphics[width=0.98\columnwidth]{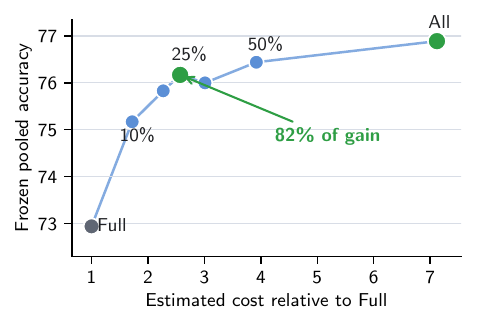}
\caption{Post-hoc Adaptive SEER budget curve on the frozen GQA-Train900 Qwen3-I/T evaluations. A label-free gate invokes complete SEER only when the Full option-score gap falls below a model-specific quantile of the unlabeled development distribution. The full curve is reported rather than selecting a quantile from frozen accuracy.}
\label{fig:sup_cost_tradeoff}
\end{figure}

Representative Qwen3-I timing on 100 GQA examples gives 0.281, 1.381, and 0.283 seconds/example for Full scoring, self-grounding, and Marker scoring. Marker, Forward, and final SEER therefore have estimated $5.92\times$, $6.91\times$, and $7.12\times$ latency, respectively; these implementation-level measurements are not hardware-independent constants. Adaptive SEER first scores Full and invokes the complete path only for low-gap examples. At unlabeled development quantiles of 10, 20, and 25\%, frozen coverage is 11.8, 20.7, and 25.7\%; estimated cost is $1.72\times$, $2.27\times$, and $2.57\times$; and pooled accuracy is 75.17, 75.83, and 76.17 from Full at 72.94. The 25\% gate gains +3.22 [1.89,4.61] with 95 fixes and 37 breaks, retaining 81.7\% of complete SEER's +3.94 gain at 36.1\% of its total cost. Because the adaptive analysis was designed after the primary protocol, it is a deployment diagnostic rather than frozen confirmation.

\begin{table}[!ht]
\centering
\small
\setlength{\tabcolsep}{2pt}
\begin{tabular*}{\columnwidth}{@{\extracolsep{\fill}}rrrrrr@{}}
\toprule
$\tau$ & Acc. & Gain & Full & Marker & Geometry \\
\midrule
0.1 & 74.38 & +1.76 & 4,056 & 12,541 & 4,175 \\
0.3 & 74.53 & +1.91 & 4,397 & 13,627 & 2,748 \\
0.5 & 74.43 & +1.81 & 4,622 & 14,589 & 1,561 \\
0.7 & 74.44 & +1.81 & 4,711 & 14,925 & 1,136 \\
\bottomrule
\end{tabular*}
\caption{Qwen3-I threshold sensitivity of the Forward ablation on 20,772 target-complete examples. Full accuracy is 72.62; Forward remains above Full for all thresholds. We use $\tau=0.5$ as a conservative global midpoint, unchanged across models and datasets.}
\label{tab:sup_threshold}
\end{table}

\section{Additional Target-Complete Results}

\subsection{Complete-Pool Relation and Decision Breakdown}

For complete-pool Qwen3-I, final SEER improves seven of nine GQA relations and six of nine VG relations. Its largest gains occur on inside (+11.86/+6.54), contains (+5.87/+9.60), and left-of (+13.46/+6.84); it loses on \emph{on} ($-4.79/-3.47$) and right-of ($-6.00/-4.27$), with a further $-0.60$ on VG behind. Qwen3-T provides a second complete-pool same-model result: Forward reaches 71.08/73.15 and final SEER 72.99/74.16 from Full at 67.10/70.28 on GQA/VG. Reciprocal consistency refinement is inactive for above/below/on by design and is most helpful for depth and containment.

Because ``contains'' is constructed by role-swapping a disjoint subset of source ``inside'' annotations, we audit whether this construction drives the pooled gain. Table \ref{tab:sup_native_audit} shows that it does not: after excluding every role-swapped example, final SEER retains +3.69 [3.26,4.11]; excluding the entire containment family as well retains +2.61 [2.19,3.04]. The larger gain on the role-swapped split indicates that containment is especially recoverable from geometry, but the effect remains on source-native relations.

\begin{table}[!ht]
\centering
\small
\setlength{\tabcolsep}{0.8pt}
\begin{tabular*}{\columnwidth}{@{\extracolsep{\fill}}lrrrr@{}}
\toprule
Split & N & Full & \method{} & Gain \\
\midrule
All & 41,544 & 70.60 & 74.60 & +3.99 \\
Source-native & 38,544 & 69.81 & 73.49 & +3.69 \\
Role-swapped & 3,000 & 80.87 & 88.77 & +7.90 \\
Native, no containment & 35,544 & 69.83 & 72.44 & +2.61 \\
\bottomrule
\end{tabular*}
\caption{Construction audit pooled over Qwen3-I/T and GQA/VG. Source-native excludes role-swapped examples; the final row also excludes both containment labels.}
\label{tab:sup_native_audit}
\end{table}

\begin{table}[!ht]
\centering
\small
\setlength{\tabcolsep}{1.2pt}
\begin{tabular*}{\columnwidth}{@{\extracolsep{\fill}}lrrrrr@{}}
\toprule
$(\kappa_F,\kappa_M)$ & N & Full & Marker & Fwd. & \method{} \\
\midrule
$(0,0)$ & 558 & 49.64 & 50.36 & 56.45 & 56.45 \\
$(0,1)$ & 2,283 & 25.71 & 74.29 & 57.73 & \textbf{74.29} \\
$(1,0)$ & 1,042 & 71.11 & 28.89 & 53.26 & \textbf{71.11} \\
$(1,1)$ & 384 & 61.20 & 38.80 & 46.09 & 46.09 \\
\midrule
All & 4,267 & 43.12 & 56.88 & 55.43 & \textbf{68.64} \\
\bottomrule
\end{tabular*}
\caption{Reciprocal-consistency structure on eligible Full/Marker conflicts. $\kappa_F$ and $\kappa_M$ indicate inverse consistency for Full and Marker. Entries after N are accuracies. The rule changes the forward source only for $(0,1)$ or $(1,0)$, which cover 3,325 cases (77.93\%).}
\label{tab:sup_routes}
\end{table}

Because Full and Marker disagree on every row in this table, exactly one of their forward predictions is correct. In the uniquely consistent rows, the consistent state is correct in 2,437 of 3,325 cases (73.29\%) and wrong in 888 (26.71\%); the latter are false-consistency cases in which an incorrect state nevertheless transforms reciprocally. Forward accuracy is 56.33\% on these rows. Reliability is highest for inside/contains (92.00/86.84\% on uniquely consistent cases) and lower for left/right (67.93/61.86\%), matching the horizontal-bias boundary reported below. Both- and neither-consistent patterns supply no state-specific evidence, so final SEER retains Forward.

\subsection{Equal-Call Reciprocal Alternatives}

All equal-call rules reuse identical forward and role-swapped Full/Marker scores. On development data, reciprocal symmetrization, averaging, and margin/entropy selection improve Forward by +1.06 to +1.44 points; unique consistency adds +1.36. A post-hoc ``any consistent, Full first'' tie rule is 0.14 points higher overall but changes sign relative to unique consistency by dataset (+0.27/+0.34 on GQA and $-0.06/-0.04$ on VG for Qwen3-I/T). Because it hard-codes a Full preference after inspection, we retain the pre-specified rule. Table~\ref{tab:sup_frozen_reciprocal_baselines} gives the complete equal-call comparison on frozen evaluations.

\begin{table*}[!ht]
\centering
\small
\setlength{\tabcolsep}{2.4pt}
\begin{tabular*}{\textwidth}{@{\extracolsep{\fill}}lrrrrr@{}}
\toprule
Equal-call rule & GQA-I & GQA-T & Emb-I & Emb-T & Emb-Intern \\
\midrule
Forward & 77.44 & 73.89 & 86.24 & 83.48 & 77.28 \\
\method{}: unique consistency & 78.44 & 75.33 & 86.24 & 83.73 & \textbf{77.53} \\
Any consistent, Full first & 78.44 & 75.33 & 86.28 & \textbf{83.81} & 77.40 \\
Any consistent, Marker first & 78.44 & 75.44 & 86.20 & 83.73 & \textbf{77.53} \\
Forward Full/Marker mean & 77.67 & 74.00 & 86.20 & 83.57 & 77.24 \\
Reciprocal Full symmetrization & 78.33 & 75.33 & 86.28 & 83.53 & 76.95 \\
Reciprocal Marker symmetrization & \textbf{78.67} & \textbf{75.67} & 86.20 & 83.65 & 77.32 \\
Reciprocal state mean & \textbf{78.67} & \textbf{75.67} & 86.28 & 83.69 & 77.36 \\
Reciprocal margin select & \textbf{78.67} & \textbf{75.67} & 86.28 & 83.73 & 77.40 \\
Reciprocal entropy select & \textbf{78.67} & \textbf{75.67} & \textbf{86.32} & 83.69 & 77.28 \\
\bottomrule
\end{tabular*}
\caption{Equal-call alternatives on frozen GQA-Train900 (900 rows/model) and EmbSpatial (2,434 rows/model). Every rule reuses identical forward and reciprocal scores. Score symmetrization requires complete inverse closure; on an inverse-supported EmbSpatial row without full closure, it retains Forward. The pre-specified SEER rule is within 0.34 points of the best rule in every column, and no alternative dominates across all five frozen evaluations.}
\label{tab:sup_frozen_reciprocal_baselines}
\end{table*}

\begin{table}[!ht]
\centering
\small
\setlength{\tabcolsep}{1.4pt}
\begin{tabular*}{\columnwidth}{@{\extracolsep{\fill}}llrrr@{}}
\toprule
Data & Model & Images & Gain & Image-cluster 95\% CI \\
\midrule
GQA & Qwen3-I & 3,234 & +3.41 & [+2.74,+4.09] \\
VG & Qwen3-I & 4,490 & +2.60 & [+1.92,+3.28] \\
GQA & Qwen3-T & 3,234 & +5.89 & [+5.12,+6.67] \\
VG & Qwen3-T & 4,490 & +3.88 & [+3.17,+4.60] \\
\midrule
All & Qwen3-I & 7,520 & +3.03 & [+2.55,+3.52] \\
All & Qwen3-T & 7,520 & +4.95 & [+4.43,+5.48] \\
\bottomrule
\end{tabular*}
\caption{Paired image-cluster bootstrap intervals for final SEER gain over Full (20,000 repetitions). Pooled rows merge identical GQA/VG numeric image IDs and retain all model outcomes per image. Every interval excludes zero.}
\label{tab:sup_cluster_ci}
\end{table}

\subsection{Self-Grounding Quality and Error Propagation}

The target-complete source records provide scene-graph boxes that are never exposed to \method{} but permit a post-hoc localization audit. Table \ref{tab:sup_grounding_quality} reports valid-output rate and mean IoU on valid outputs. Self-grounding is usually parseable, but its boxes are noisy and model dependent; this is why \method{} treats localization as evidence to assess rather than as a trusted detector output.

\begin{table}[!ht]
\centering
\small
\setlength{\tabcolsep}{2pt}
\begin{tabular*}{\columnwidth}{@{\extracolsep{\fill}}llrrrr@{}}
\toprule
Data & Grounder & N & Valid & S IoU & O IoU \\
\midrule
GQA & Qwen3-I & 11,070 & 99.35 & 41.87 & 43.51 \\
VG & Qwen3-I & 9,702 & 99.36 & 32.79 & 45.42 \\
GQA & Qwen3-T & 11,070 & 98.72 & 37.93 & 38.89 \\
VG & Qwen3-T & 9,702 & 98.31 & 28.83 & 41.09 \\
\bottomrule
\end{tabular*}
\caption{Target-relation-hidden self-grounding audit. Valid, S IoU, and O IoU are percentages; IoU is averaged over valid outputs against source scene-graph boxes and is not available to the decision rule.}
\label{tab:sup_grounding_quality}
\end{table}

Table \ref{tab:sup_iou_propagation} pools Qwen3-I/T Forward predictions by the smaller subject/object IoU. The gain is exactly zero for invalid outputs because they are forced to Full. Gains increase to +2.89--+4.74 once both boxes have nonzero overlap with the annotations. The zero-IoU group remains slightly positive, which can occur when the model localizes a visually relevant instance that differs from the scene-graph instance; therefore IoU is diagnostic rather than a test-time gate. Reciprocal decision quality is reported separately in Table~\ref{tab:sup_routes}.

\begin{table}[!ht]
\centering
\small
\setlength{\tabcolsep}{2pt}
\begin{tabular*}{\columnwidth}{@{\extracolsep{\fill}}lrrrrr@{}}
\toprule
Min S/O IoU & N & Full & Marker & Forward & Gain \\
\midrule
Invalid & 440 & 68.41 & 68.64 & 68.41 & +0.00 \\
0 & 15,182 & 64.48 & 63.47 & 64.95 & +0.47 \\
$(0,.10)$ & 6,513 & 70.90 & 74.11 & 73.79 & +2.89 \\
$[.10,.30)$ & 6,646 & 72.30 & 77.31 & 77.04 & +4.74 \\
$\geq .30$ & 12,763 & 76.93 & 80.98 & 81.00 & +4.07 \\
\bottomrule
\end{tabular*}
\caption{Forward grounding-quality propagation over 41,544 target-complete model--example evaluations. Source boxes are used only for this post-hoc analysis.}
\label{tab:sup_iou_propagation}
\end{table}

The state inventory also gives a direct Forward transition breakdown. Marker is selected in 2,377 cases where Full is wrong and Marker is correct; Full is retained in 1,436 cases where Full is correct and Marker is wrong. Geometry is the only correct candidate in 157 cases where both visual states fail, but is wrong in 169 cases where at least one visual state is correct. These candidate-level counts differ from the direct No-Geometry routing ablation above because the larger-gap visual rule does not always select the correct available visual state. Together, the analyses show why neither ``always use Marker'' nor ``always use Geometry'' is sufficient: role-conditioned evidence supplies most repairs, while abstention and Full fallback contain its failures.

\subsection{Independent Human Audit}

Two annotators independently inspected 240 valid predicted-grounding cards, sampling 15 items from each dataset--grounder--IoU stratum, without model, dataset, IoU, or population-stratum metadata. Separately, they inspected 270 source-pair cards, sampling 15 items for each dataset and relation, without gold labels. The first audit asks whether the predicted S/O regions bind the intended entities and leave their local relation judgeable; the second checks whether the source pair is valid, the displayed relation is visually answerable, and the candidate answer is unique. Both annotators were university peers of the authors. They volunteered without compensation and gave informed consent after receiving a complete description of the annotation task and its intended research use. The audit collected only task judgments; no demographic, identifying, or sensitive information was requested. Table~\ref{tab:sup_human_audit} reports sampling-weighted percentages for the valid-grounding population and unweighted percentages for the relation-balanced semantic audit. Agreement and Cohen's $\kappa$ are computed on the audited items before population weighting.

\begin{table}[!ht]
\centering
\small
\setlength{\tabcolsep}{1.4pt}
\begin{tabular*}{\columnwidth}{@{\extracolsep{\fill}}lrrrrr@{}}
\toprule
Audit decision & N & Ann. 1 & Ann. 2 & Agree & $\kappa$ \\
\midrule
Grounding pair correct & 240 & 74.80 & 79.23 & 89.17 & .657 \\
Local relation judgeable & 240 & 81.63 & 82.40 & 88.75 & .564 \\
Source pair correct & 270 & 95.56 & 94.07 & 95.56 & .551 \\
Relation answerable & 270 & 92.59 & 92.22 & 90.74 & .344 \\
Best answer is gold & 270 & 78.15 & 80.00 & 83.70 & .749 \\
Another option is valid & 270 & 11.11 & 10.00 & 96.30 & .814 \\
\bottomrule
\end{tabular*}
\caption{Independent human-audit results (\%). For the best-answer row, annotator columns measure agreement with the hidden gold answer, whereas inter-annotator agreement compares their exact selected options. Modest binary $\kappa$ values can coexist with high raw agreement because positive judgments dominate pair correctness and answerability.}
\label{tab:sup_human_audit}
\end{table}

We next form a strict consensus subset: both annotators must mark the source pair correct and the relation answerable, choose the hidden gold answer, and reject any second valid option. This retains 180 of 270 items. Table~\ref{tab:sup_human_clean} links these items to the unchanged Qwen3-I/T predictions. The pooled +4.72-point gain remains positive under an image-cluster bootstrap (95\% CI [+0.83,+8.61]), so visibly invalid or multiply answerable source annotations do not fully explain the primary effect. This is a post-hoc validity analysis, not a replacement benchmark estimate. The filter is intentionally conservative and leaves only 5 of the 30 audited \emph{on} examples, where contact and overlap often make more than one displayed relation defensible.

\begin{table}[!ht]
\centering
\small
\setlength{\tabcolsep}{2pt}
\begin{tabular*}{\columnwidth}{@{\extracolsep{\fill}}lrrrrrr@{}}
\toprule
Model & N & Full & Marker & Forward & Gain & Fix/Break \\
\midrule
Qwen3-I & 180 & 73.89 & 78.89 & 77.78 & +3.89 & 15/8 \\
Qwen3-T & 180 & 64.44 & 72.22 & 70.00 & +5.56 & 16/6 \\
\midrule
Pooled & 360 & 69.17 & 75.56 & 73.89 & +4.72 & 31/14 \\
\bottomrule
\end{tabular*}
\caption{Forward performance on the strict two-annotator semantic-consensus subset. N counts model--example evaluations; Fix/Break compares Forward with Full.}
\label{tab:sup_human_clean}
\end{table}

To audit the frozen set directly, the same annotators later evaluated 120 unique-image rows sampled before inspection: 30 target-\emph{on} rows, 30 horizontal rows with repeated entity names, 30 horizontal rows with unique names, and 30 other repeated-name rows. The two-stage interface first shows the original image and entity names without boxes, then reveals the source S/O boxes; gold labels, model outputs, method routes, and sampling-stratum names remain hidden. Table~\ref{tab:sup_frozen_human_strata} separates referential ambiguity in the unboxed input from source-pair validity after the intended instances are shown. Aggregate raw agreement/Cohen's $\kappa$ is 85.83/.702 for unboxed pair uniqueness, 72.50/.645 for the unboxed answer, 89.17/.391 for source-pair correctness, 96.67/.944 for the boxed answer, and 97.50/.904 for whether another option is valid. The low unboxed uniqueness rate confirms that repeated instances make this a genuinely binding-sensitive subset; the high boxed-gold agreement shows that most rows become semantically clear once the intended pair is exposed.

\begin{table*}[!tb]
\centering
\small
\setlength{\tabcolsep}{2.5pt}
\begin{tabular*}{\textwidth}{@{\extracolsep{\fill}}lrrrrrr@{}}
\toprule
Frozen audit stratum & N & Unboxed unique A/B & Source pair A/B & Boxed gold A/B & Multi-valid A/B & Strict \\
\midrule
Horizontal, repeated name & 30 & 13.33/16.67 & 86.67/73.33 & 100.00/93.33 & 0.00/0.00 & 21 \\
Horizontal, unique names & 30 & 33.33/46.67 & 83.33/90.00 & 100.00/100.00 & 0.00/0.00 & 23 \\
Other repeated name & 30 & 36.67/40.00 & 96.67/96.67 & 90.00/90.00 & 3.33/3.33 & 25 \\
Target \emph{on} & 30 & 63.33/60.00 & 100.00/96.67 & 93.33/93.33 & 63.33/53.33 & 9 \\
\midrule
All & 120 & 36.67/40.83 & 91.67/89.17 & 95.83/94.17 & 16.67/14.17 & 78 \\
\bottomrule
\end{tabular*}
\caption{Targeted two-annotator audit of frozen GQA-Train900 (\%). ``Unboxed unique'' asks whether the named pair is uniquely identifiable before boxes are shown. ``Strict'' requires both annotators to accept the boxed source pair and answerability, select the hidden gold, and reject a second valid option. The strata deliberately stress horizontal order, repeated instances, and \emph{on}; they are not a prevalence-weighted estimate of all 900 rows.}
\label{tab:sup_frozen_human_strata}
\end{table*}

The complete frozen set also permits a model-output-independent name-multiplicity split: 535 images contain exactly one scene-graph instance of each named entity, while 365 contain a repeated subject or object name. Table~\ref{tab:sup_frozen_human_clean} shows that both groups retain positive pooled gains. The targeted audit further separates the 78 strict semantic-consensus rows by whether both annotators could identify the named pair before boxes were shown. The 23 unanimously unique images retain +8.70 [2.17,17.39], whereas the 47 unanimously non-unique images retain +6.38 [$-4.26$,17.02]; eight disagreements are included only in the all-strict row. Thus gains are not confined to repeated names or manually identified ambiguous pairs. Across all 78 strict rows, the per-model gains are +7.69 for Qwen3-I and +3.85 for Qwen3-T, pooling to +5.77 [$-1.28$,13.46]. Because the audit was deliberately targeted and its groups are small, these rows diagnose ambiguity rather than estimate its population prevalence.

\begin{table}[!tb]
\centering
\small
\setlength{\tabcolsep}{0.6pt}
\begin{tabular*}{\columnwidth}{@{\extracolsep{\fill}}lrrrrrr@{}}
\toprule
Slice & N & Full & Marker & \method{} & Gain & $\mathrm{CI}_{img}$ \\
\midrule
\multicolumn{7}{l}{\emph{Complete frozen set}} \\
Unique names & 535 & 75.42 & 77.38 & 77.66 & +2.24 & [0.37,4.11] \\
Repeated name & 365 & 69.32 & 73.15 & 72.74 & +3.42 & [0.41,6.44] \\
\midrule
\multicolumn{7}{l}{\emph{Targeted strict human audit}} \\
Unboxed unique & 23 & 65.22 & 78.26 & 73.91 & +8.70 & [2.17,17.39] \\
Unboxed non-unique & 47 & 71.28 & 73.40 & 77.66 & +6.38 & [$-4.26$,17.02] \\
All strict & 78 & 66.67 & 71.79 & 72.44 & +5.77 & [$-1.28$,13.46] \\
\bottomrule
\end{tabular*}
\caption{Referential-ambiguity slices on frozen GQA-Train900. Scores pool Qwen3-I/T; N counts images, and intervals cluster both model outcomes by image. Name multiplicity is computed on all 900 frozen scene graphs. Human-audit rows require strict semantic consensus; ``unboxed'' records whether both annotators identified the named pair before source boxes were shown.}
\label{tab:sup_frozen_human_clean}
\end{table}

Because both human audits flag \emph{on} as the least clean target, Table~\ref{tab:sup_no_on} recomputes the complete automatic evaluations under two relation-defined, model-output-independent reporting slices. Removing only \emph{on} targets preserves the other eight target labels. Removing every candidate set that contains \emph{on} is stricter and drops the complete above/below/on family. Both filters increase the SEER--Full gain on the frozen and development pools; semantic ambiguity around \emph{on} therefore cannot account for the aggregate improvement.

\begin{table*}[!ht]
\centering
\small
\setlength{\tabcolsep}{2.5pt}
\begin{tabular*}{\textwidth}{@{\extracolsep{\fill}}llrrrrrr@{}}
\toprule
Pool & Reporting slice & N & Full & Marker & Fwd. & \method{} & $\Delta$Full \\
\midrule
Frozen GQA & Exclude target \emph{on} & 1,600 & 72.25 & 76.12 & 75.62 & \textbf{77.00} & +4.75 \\
Frozen GQA & Exclude \emph{on} option family & 1,200 & 75.83 & 79.17 & 79.00 & \textbf{80.83} & +5.00 \\
Development GQA/VG & Exclude target \emph{on} & 36,164 & 69.51 & 73.24 & 73.13 & \textbf{74.69} & +5.19 \\
Development GQA/VG & Exclude \emph{on} option family & 25,404 & 71.77 & 74.89 & 75.28 & \textbf{77.50} & +5.74 \\
\bottomrule
\end{tabular*}
\caption{Robustness to \emph{on}-related semantic ambiguity. N counts model--example evaluations. The option-family slice removes every above/below/on candidate set, not only rows whose target is \emph{on}.}
\label{tab:sup_no_on}
\end{table*}

Exact source-instance recovery is not required for a benchmark answer to improve. After weighting the 240 grounding cards to the valid target-complete population, Full/Marker/Forward accuracy is 70.59/72.14/72.89 when both annotators accept the grounding, 65.52/78.21/78.68 when both reject it, and 57.39/66.66/71.26 when they disagree. The larger gains in the latter strata can arise from same-class or contextual regions, scene context, and answer priors. Consequently, the human audit supports inspectable, query-conditioned evidence construction and quantifies its localization errors; it does not make benchmark gain a certificate of exact reference faithfulness.

\subsection{Target-Complete Mechanism Controls}

We isolate the two operations inside the marker state on the balanced target-complete sets. Crop uses the target-relation-hidden self-grounded union without visible boxes or role labels. Self-Marker adds the S/O overlay using exactly the same predicted boxes. Oracle-Marker replaces those boxes with source scene-graph annotations and is a diagnostic upper bound, not a deployable \method{} setting. For Swapped and Random, the self-grounded crop is held fixed: Swapped exchanges the S/O marker identities, while Random moves same-size markers to deterministic positions within the crop. Invalid self-grounding retains the original image in all self-grounded conditions. Thus the counterfactuals corrupt S/O role assignment without changing the question, candidates, evaluated VLM, or scoring rule.

\begin{table}[!ht]
\centering
\small
\setlength{\tabcolsep}{1.4pt}
\begin{tabular*}{\columnwidth}{@{\extracolsep{\fill}}llrrrrrr@{}}
\toprule
Model & Data & Full & Crop & Self-M & Oracle-M & Swap & Random \\
\midrule
Qwen3-I & GQA & 75.00 & 75.44 & 76.67 & 84.00 & 75.56 & 75.33 \\
Qwen3-I & VG & 70.33 & 72.00 & 72.11 & 77.22 & 72.00 & 68.78 \\
Qwen3-T & GQA & 69.11 & 70.78 & 74.00 & 80.22 & 71.67 & 72.00 \\
Qwen3-T & VG & 66.67 & 67.89 & 69.89 & 75.89 & 66.78 & 66.56 \\
\midrule
Pooled & Both & 70.28 & 71.53 & 73.17 & 79.33 & 71.50 & 70.67 \\
\bottomrule
\end{tabular*}
\caption{Matched mechanism controls over 3,600 target-complete model--example evaluations. Oracle-M is an annotation-box upper bound. Swap and Random retain the self-grounded crop but corrupt S/O role assignment.}
\label{tab:sup_targetcomplete_mechanism}
\end{table}

The pooled ladder separates local refocus (Full$\rightarrow$Crop, +1.25) from role-explicit conditioning (Crop$\rightarrow$Self-Marker, +1.64). The latter remains positive under underlying-image-clustered bootstrap, with 95\% interval [+0.25,+3.05]; this clustering keeps cross-dataset and cross-model outcomes for one image together. Oracle-Marker adds a further +6.17, exposing substantial headroom from localization quality, while Swapped and Random reduce Self-Marker by 1.67 and 2.50 points. Tuple-level exact McNemar tests give $p<.001$ for the oracle and random comparisons and $p=.0054$ for Swapped, but are treated only as descriptive because images and VLMs repeat. The same qualitative ordering holds in all four rows. The larger marker and counterfactual effects for the Qwen3-T checkpoint show that its predictions remain sensitive to which visual entities are assigned to the query roles; they do not isolate free-form deliberation.

\begin{table}[!ht]
\centering
\small
\setlength{\tabcolsep}{2pt}
\begin{tabular*}{\columnwidth}{@{\extracolsep{\fill}}lrrr@{}}
\toprule
Frozen Qwen3-I condition & Acc. & Gain & Fix/Break \\
\midrule
Full / direct & 75.33 & -- & -- \\
Full / role text & 75.44 & +0.11 & 18/17 \\
Full / S/O marker & \textbf{76.89} & +1.56 & 72/58 \\
Crop / direct & 75.89 & +0.56 & 32/27 \\
Crop / Marker text, no marks & 75.33 & 0.00 & 39/39 \\
Crop / role text & 76.67 & +1.33 & 40/28 \\
Crop / unlabeled red--blue boxes & 76.67 & +1.33 & 78/66 \\
Crop / labeled S/O marker & 76.11 & +0.78 & 75/68 \\
\bottomrule
\end{tabular*}
\caption{Matched interface controls on frozen GQA-Train900 using the same self-grounded boxes. The family separates image extent, role wording, visible box locations, and S/O glyphs; visual conditions use the corresponding box/marker instruction.}
\label{tab:sup_frozen_marker_controls}
\end{table}

The identical Marker instruction without visible marks returns exactly to Full accuracy despite changing 78 predictions, and role text on Full changes almost nothing. Visible query locations therefore contribute beyond extra wording. Full-image markers outperform their cropped counterpart, showing that crop context loss can offset local refocus; unlabeled color boxes also match role-text crops and exceed labeled S/O markers. We accordingly attribute the intervention to an explicit location-and-role interface, not uniquely to the glyphs or crop.

We additionally expand Crop and Oracle-Marker to all 20,772 complete-pool Qwen3-I examples. On GQA, Full/Crop/Self-Marker/Forward/SEER/Oracle-Marker obtain 72.40/73.76/74.03/74.27/75.82/82.34; on VG they obtain 72.87/74.49/73.76/74.60/75.47/78.29. Crop improves both datasets, but Self-Marker does not uniformly dominate it: +0.27 on GQA and $-0.73$ on VG. In the pooled result, Forward exceeds both single transformed states and final SEER adds +1.23, matching the intended setting in which the useful transformation varies by example.

Paired Crop-versus-Full gains are +1.36 on GQA and +1.62 on VG, with image-cluster 95\% intervals [+0.77,+1.94] and [+1.09,+2.16]. Self-Marker versus Crop remains uncertain: +0.27 [-0.47,+1.03] on GQA and $-0.73$ [-1.50,+0.04] on VG. Oracle-Marker exceeds Self-Marker by +8.31 [6.60,10.06] on GQA and +4.54 [3.76,5.31] on VG, giving a pooled +6.55-point grounding headroom. Bracketed quantities are paired image-cluster 95\% intervals.

\subsection{Balanced Cross-Model Paired Statistics}

The main paper reports the symmetric four-model/two-dataset matrix under both grounding-order protocols; Table~\ref{tab:sup_paired_balanced} adds paired changes for the pre-specified subject-first rows and a separate InternVL3.5 GQA replication. Every row has more corrections than regressions, while the two smallest effects are Qwen3-I and R1-OV on VG. The pooled matrix is descriptive and does not treat repeated images or models as independent.

\begin{table*}[!tb]
\centering
\small
\setlength{\tabcolsep}{2.5pt}
\begin{tabular*}{\textwidth}{@{\extracolsep{\fill}}lllrrrrr@{}}
\toprule
Data & VLM & $G$ & Full & \method{} & Gain & F/B & $p$ \\
\midrule
GQA & Qwen3-I & Same & 75.00 & 78.67 & +3.67 & 69/36 & .0017 \\
GQA & Qwen3-T & Same & 69.11 & 75.67 & +6.56 & 86/27 & $2.3{\times}10^{-8}$ \\
GQA & Qwen2.5-I & Q3-I & 73.11 & 78.22 & +5.11 & 67/21 & $9.2{\times}10^{-7}$ \\
GQA & R1-OV & Q3-I & 65.78 & 70.89 & +5.11 & 74/28 & $5.9{\times}10^{-6}$ \\
GQA & InternVL3.5 & Same & 61.89 & 70.78 & +8.89 & 130/50 & $2.1{\times}10^{-9}$ \\
\midrule
VG & Qwen3-I & Same & 70.33 & 72.67 & +2.33 & 66/45 & .057 \\
VG & Qwen3-T & Same & 66.67 & 71.56 & +4.89 & 89/45 & .00018 \\
VG & Qwen2.5-I & Q3-I & 68.89 & 72.56 & +3.67 & 74/41 & .0027 \\
VG & R1-OV & Q3-I & 62.67 & 64.11 & +1.44 & 48/35 & .188 \\
\bottomrule
\end{tabular*}
\caption{Balanced paired statistics. $G$ is the grounder and F/B is fix/break. InternVL3.5 is an additional same-model GQA replication; no adjusted row-level claim is made.}
\label{tab:sup_paired_balanced}
\end{table*}

InternVL3.5 has 690 image clusters. Marker alone gains +2.00, Forward gains +6.89, and final SEER gains +8.89, with 130 fixes against 50 breaks relative to Full (exact McNemar $p=2.1\times10^{-9}$). Reciprocal consistency refinement contributes +2.00 over Forward with 39 fixes against 21 breaks ($p=.0273$, image-cluster interval [+0.22,+3.86]). Target-relation-hidden same-model grounding succeeds on 883/900 examples (98.11\%), with mean subject/object IoU 0.458/0.469 against source boxes. This result extends same-model grounding beyond the Qwen architecture while retaining the same evidence states and decision rule.

\subsection{SEER-Open: Option-Free Generation}

\method{}-Open is an option-free adaptation rather than the main paper's closed-set equation applied unchanged. Its prompt hides candidate relations and asks for one short spatial phrase. A deterministic parser maps common surface forms such as ``to the left of,'' ``contained in,'' and ``in front of'' to nine canonical labels, with unparsable responses counted as wrong. Geometry retains the same eligibility rule; when it abstains, \method{}-Open prefers the only parseable state if exactly one is parseable, otherwise uses the higher mean generated-token log-probability, and falls back to Full on an exact tie. Paired image-cluster 95\% intervals for its gain over Full are [+2.26,+11.17] on GQA and [-0.56,+10.61] on VG; the smaller VG sample consequently yields the wider interval.

\begin{table}[!tb]
\centering
\small
\setlength{\tabcolsep}{2pt}
\begin{tabular*}{\columnwidth}{@{\extracolsep{\fill}}lrrrr@{}}
\toprule
Data & N & Full & Marker & SEER-O \\
\midrule
GQA & 180 & 67.78 & 70.00 & 74.44 \\
VG & 180 & 63.89 & 63.89 & 68.89 \\
\bottomrule
\end{tabular*}
\caption{Option-free accuracy with 20 examples per relation. SEER-O denotes \method{}-Open.}
\label{tab:sup_open_relation}
\end{table}

\section{Matched Component and Tool Controls}

The component controls in Table~\ref{tab:sup_component_controls} use independently constructed relation-balanced GQA900 and VG630 diagnostics. Their candidate construction differs from the target-complete protocol, so they are not pooled with the primary tables. They remain useful for component attribution because every row compares identical sample identifiers, answer options, option order, and VLM scoring; only the evidence interface changes. Subsequent external-tool comparisons use the subsets stated in their captions and are reported separately.

\begin{table*}[!t]
\centering
\footnotesize
\setlength{\tabcolsep}{3pt}
\begin{tabularx}{\textwidth}{@{}>{\raggedright\arraybackslash}p{0.15\textwidth}>{\raggedright\arraybackslash}p{0.14\textwidth}>{\raggedright\arraybackslash}p{0.12\textwidth}>{\raggedright\arraybackslash}p{0.10\textwidth}>{\raggedright\arraybackslash}p{0.12\textwidth}>{\raggedright\arraybackslash}X@{}}
\toprule
Interface & Regions & Query scope & Target hidden & Role-explicit & Alternatives / decision \\
\midrule
Query crop/self-grounding & Varies & Query region & Not required & Usually implicit & One transformed view \\
Set-of-Mark \citep{yang2023som} & External segmentation & Dense scene & N/A & Numeric marks & One marked view \\
Graph-of-Mark \citep{frisoni2026graphofmark} & Det./seg./depth graph & Question-filtered & No & Graph nodes/edges & One graph-enriched view \\
\method{} & Frozen VLM grounder & Queried S/O pair & Yes & Explicit S/O & Complementary evidence; fixed forward/reciprocal rule \\
\bottomrule
\end{tabularx}
\caption{Protocol-level component comparison. Individual primitives such as cropping, marks, or self-grounding are not claimed as new; \method{} contributes their constrained conjunction: pair-specific target-relation-hidden localization, semantic-role conditioning, alternative evidence states, and a fixed relation-aware decision around an unchanged VLM.}
\label{tab:sup_protocol_components}
\end{table*}

\subsection{Crop, Prompt Recipe, and Text Coordinates}

Crop reuses the target-relation-hidden boxes but shows only the padded subject/object union, without boxes or role labels. SOP gives the model a textual sequence---localize the entities, estimate boxes, compare them, and answer---without materializing an intermediate evidence view. Text-BBox exposes the same predicted coordinates as text over the original image. Table \ref{tab:sup_component_controls} shows that crop-only recovers a modest +2.06 points overall and explicit S/O markers do slightly better, whereas merely describing the procedure or serializing noisy coordinates into text performs poorly under the matched scoring interface. The result supports role-explicit visual conditioning rather than additional instructions as the useful intervention.

\begin{table*}[!tb]
\centering
\small
\setlength{\tabcolsep}{3pt}
\begin{tabular*}{\textwidth}{@{\extracolsep{\fill}}llrrrrrr@{}}
\toprule
Data & VLM & N & Full & Crop & S/O Marker & SOP & Text-BBox \\
\midrule
GQA900 & Qwen3-I & 900 & 66.78 & 68.78 & 70.00 & 53.22 & 52.89 \\
GQA900 & Qwen3-T & 900 & 62.44 & 65.11 & 67.11 & 48.67 & 46.78 \\
VG630 & Qwen3-I & 630 & 67.78 & 68.57 & 66.67 & 52.86 & 53.81 \\
VG630 & Qwen3-T & 630 & 65.56 & 68.10 & 66.98 & 48.25 & 45.08 \\
\midrule
All & Sample weighted & 3,060 & 65.46 & 67.52 & 67.84 & 50.78 & 49.67 \\
\bottomrule
\end{tabular*}
\caption{Matched evidence-interface controls on the relation-balanced diagnostic sets. These rows isolate components and are not target-complete benchmark results.}
\label{tab:sup_component_controls}
\end{table*}

A latency-matched control uses seven genuinely distinct Full-image prompts, with neither grounding nor transformed visual evidence. On the balanced GQA/VG sets, soft score averaging obtains 74.44/70.89 for Qwen3-I and 69.22/66.44 for Qwen3-T; majority voting obtains 74.78/70.89 and 68.89/66.56. The corresponding final SEER results are 78.67/72.67 and 75.67/71.56. Across all 3,600 model--example rows, soft and vote ensembles obtain 70.25 and 70.28 versus 74.64 for SEER; the descriptive paired differences are +4.39 (317/159 fixes/breaks) and +4.36 points (314/157). Seven Full scores cost $7.00\times$ one Full score, closely matching the measured $7.12\times$ final SEER path, so repeated full-image prompting at the same budget does not recover the evidence-interface gain.

\subsection{External Grounders and Released Marking Pipelines}

We compare pair-specific evidence with two external-tool settings. First, OWL-ViT \citep{minderer2022owlvit} receives only the subject and object phrases. On its detector-success subsets, rendering the two detected boxes as S/O markers improves Qwen3-I from 67.01 to 69.98 on 876 GQA examples and from 67.91 to 68.72 on 617 VG examples, whereas deterministic geometry from those boxes obtains 45.78 and 44.41. This diagnostic shows that external localization can support the interface, but box geometry alone does not resolve all nine relations.

Second, Table~\ref{tab:sup_external_marks} evaluates the official released high-level Graph-of-Mark API \citep{frisoni2026graphofmark} on every example of the balanced GQA900 and VG900 sets. We use repository commit \texttt{5d9ea6e}, its native question-conditioned filtering, YOLOv8x, SAM-HQ ViT-B, Depth Anything V2 Large, and no CLIP stage. SoM is the released Set-of-Mark rendering \citep{yang2023som}; GoM adds the released graph/depth visualization, and GoM+labels also displays textual relation labels. The original multiple-choice question is supplied for native filtering, but the gold answer is never exposed. Every visual input is scored by the same VLM with the same ordered options. This public API differs from the Graph-of-Mark paper's full three-detector configuration, so the experiment is a reproducible released-pipeline comparison rather than an exact paper reproduction.

\begin{table}[!ht]
\centering
\small
\setlength{\tabcolsep}{1pt}
\begin{tabular*}{\columnwidth}{@{\extracolsep{\fill}}llrrrrr@{}}
\toprule
Data & VLM & Full & SoM & GoM & +Label & \method{} \\
\midrule
GQA & Qwen2.5-I & 73.11 & 71.56 & 71.78 & 69.56 & \textbf{78.22} \\
VG & Qwen2.5-I & 68.89 & 64.00 & 63.89 & 63.22 & \textbf{72.56} \\
GQA & Qwen3-I & 75.00 & 71.11 & 72.33 & 70.00 & \textbf{78.67} \\
VG & Qwen3-I & 70.33 & 68.44 & 67.44 & 66.00 & \textbf{72.67} \\
\bottomrule
\end{tabular*}
\caption{Matched comparison with the official released Graph-of-Mark high-level API. Every row contains 900 examples and uses identical closed-set option scoring. ``+Label'' is GoM with textual relation labels. Qwen3-I uses same-model self-grounding for \method{}; Qwen2.5-I uses the fixed Qwen3-I grounder used throughout the balanced evaluation.}
\label{tab:sup_external_marks}
\end{table}

Dense public mark renderings do not improve on Full in these rows, whereas \method{}'s pair-specific evidence interface improves both VLMs. The result supports task--interface matching, not a general claim that \method{} supersedes the full Graph-of-Mark system.

\section{Mechanism and Transfer Diagnostics}

\subsection{COCO-Geom}

COCO-Geom is a controlled 400-example diagnostic derived from COCO instance boxes \citep{lin2014coco}. Its labels intentionally follow 2D geometry, so it tests whether target-relation-hidden VLM localization can recover a box-derived decision rather than general human spatial semantics. The set includes diagonal distractors, overlap, small margins, and non-horizontal relations. GT-Box View renders the source COCO boxes for the VLM and is an analysis-only visual upper bound; Blind Geometry and Box+Geometry instead use target-relation-hidden self-grounded boxes. Because GT-Box View still asks the VLM to classify the marked image, it is distinct from deterministic geometry computed directly from source boxes. The latter reaches 100\% by construction and is therefore a label-generation ceiling rather than a model result.

\begin{table}[!ht]
\centering
\small
\setlength{\tabcolsep}{1pt}
\begin{tabular*}{\columnwidth}{@{\extracolsep{\fill}}lrrrrr@{}}
\toprule
Model & Full & GT-Box V. & Blind G. & Box & Box+G. \\
\midrule
Qwen3-T & 55.75 & 74.75 & 84.50 & 64.50 & 83.50 \\
Qwen3-I & 63.50 & 81.00 & 84.75 & 70.50 & 85.25 \\
R1-OV & 41.75 & 48.25 & 83.50 & 43.75 & 83.50 \\
Qwen2.5-I & 56.25 & 73.00 & 84.00 & 64.00 & 86.00 \\
\bottomrule
\end{tabular*}
\caption{COCO-Geom mechanism diagnostic. These box-derived labels favor geometric evidence by construction.}
\label{tab:sup_coco}
\end{table}

\subsection{ARO VG-Relation}

ARO VG-Relation \citep{yuksekgonul2023bags} evaluates caption pairs whose relation differs. The balanced 600-example diagnostic separates local refocus from role-explicit marker conditioning across seven models. Self-grounded crop-only is nearly neutral, while visible boxes and S/O markers yield larger mean gains. Because both captions contain relation phrases, this is answer-blind caption-pair grounding rather than the cleaner target-relation-hidden protocol used for GQA/VG.

\begin{table}[!ht]
\centering
\small
\setlength{\tabcolsep}{1pt}
\begin{tabular*}{\columnwidth}{@{\extracolsep{\fill}}lrrrrr@{}}
\toprule
Model & Full & Crop & Box & S/O Marker & Marker gain \\
\midrule
Qwen3-T & 79.33 & 79.83 & 83.50 & 86.00 & +6.67 \\
Qwen3-I & 79.17 & 80.17 & 83.83 & 88.17 & +9.00 \\
Qwen3-2B & 73.33 & 73.50 & 79.67 & 82.50 & +9.17 \\
R1-OV & 75.17 & 75.17 & 83.17 & 84.50 & +9.33 \\
Qwen2.5-I & 79.50 & 79.67 & 83.33 & 86.17 & +6.67 \\
LLaVA-1.5 & 59.67 & 59.33 & 62.67 & 59.17 & -0.50 \\
InternVL3.5 & 65.83 & 65.17 & 67.50 & 79.17 & +13.33 \\
\midrule
Mean & 73.14 & 73.26 & 77.67 & 80.81 & +7.67 \\
\bottomrule
\end{tabular*}
\caption{ARO evidence-view ablation. Crop isolates local refocus; Box adds explicit locations; S/O Marker adds role identity.}
\label{tab:sup_aro}
\end{table}

On the full official 23,937-example ARO VG-Relation set, Qwen3-I improves from 82.51 on the original image to 87.08 using the ARO-provided relation-region crop (+4.57). This uses ARO metadata rather than self-grounding, so it supports the localization diagnosis but is not a deployable \method{} result.

\subsection{Frozen EmbSpatial Pair-Relation Test}

EmbSpatial-Bench \citep{du2024embspatial} contains spatial multiple-choice questions from AI2-THOR, Matterport3D, and ScanNet. We retain all 2,434 official test rows with exactly two queried objects and a left, right, above, or under target. The official question, options, order, and answer remain unchanged. Qwen3-I, Qwen3-T, and InternVL3.5 each perform target-relation-hidden same-model grounding and do not receive the answer options during localization. Official boxes are excluded from Full scoring, grounding, Marker construction, and routing; they are used only for the post-hoc IoU diagnostic below. The decision protocol was fixed before Qwen3-I evaluation and left unchanged for the subsequent replications.

\begin{table}[!ht]
\centering
\small
\setlength{\tabcolsep}{1.0pt}
\begin{tabular*}{\columnwidth}{@{\extracolsep{\fill}}llrrrr@{}}
\toprule
Model & Source & Full & Marker & \method{} & Gain \\
\midrule
Qwen3-I & AI2-THOR & 79.23 & 84.78 & \textbf{85.87} & +6.64 \\
& MP3D & 83.13 & 84.00 & \textbf{84.62} & +1.49 \\
& ScanNet & 83.38 & 87.25 & \textbf{88.25} & +4.87 \\
& All & 81.88 & 85.33 & \textbf{86.24} & +4.35 \\
\midrule
Qwen3-T & AI2-THOR & 77.54 & 81.52 & \textbf{82.85} & +5.31 \\
& MP3D & 79.16 & 83.50 & \textbf{83.75} & +4.59 \\
& ScanNet & 79.25 & 83.88 & \textbf{84.62} & +5.37 \\
& All & 78.64 & 82.95 & \textbf{83.73} & +5.09 \\
\midrule
InternVL3.5 & AI2-THOR & 65.70 & 76.21 & \textbf{77.05} & +11.35 \\
& MP3D & 62.78 & \textbf{78.41} & 76.30 & +13.52 \\
& ScanNet & 68.75 & \textbf{79.62} & 79.25 & +10.50 \\
& All & 65.74 & \textbf{78.06} & 77.53 & +11.79 \\
\bottomrule
\end{tabular*}
\caption{Frozen-protocol primary SEER results on all official EmbSpatial pair-relation test rows. Sources contain 828/86, 806/26, and 800/164 examples/scenes for AI2-THOR, Matterport3D (MP3D), and ScanNet, totaling 2,434 examples from 276 scenes. All-source scene-cluster 95\% intervals for Qwen3-I, Qwen3-T, and InternVL3.5 are [+3.12,+5.68], [+3.81,+6.47], and [+10.14,+13.36].}
\label{tab:sup_embspatial}
\end{table}

For Qwen3-I, SEER produces 170 fixes and 64 breaks relative to Full (exact McNemar $p=2.83\times10^{-12}$); gains by target relation are +6.88 for above, +8.47 for below, +3.90 for left, and $-1.61$ for right. Grounding succeeds on 2,426/2,434 examples (99.67\%), with mean subject/object IoU 0.367/0.356 against withheld official boxes. Reciprocal consistency refinement is eligible on 31 examples and has one fix and one break relative to Forward, hence zero net effect.

For Qwen3-T, SEER produces 186 fixes and 62 breaks ($p=1.42\times10^{-15}$); relation gains are +7.72 for above, +12.46 for below, +1.62 for left, and $-1.13$ for right. Grounding succeeds on 2,405/2,434 examples (98.81\%), with mean IoU 0.339/0.326. Reciprocal consistency refinement is eligible on 39 examples and adds 0.25 points over Forward.

InternVL3.5 yields 364 fixes and 77 breaks ($p=1.11\times10^{-45}$), with relation gains of +18.62/+13.79/+6.66/+8.39 for above/below/left/right. Grounding succeeds on 2,385/2,434 examples (97.99\%; mean IoU 0.311/0.365). Among 65 eligible cases, reciprocal consistency refinement makes six fixes and no breaks, adding 0.25 points over Forward. Marker remains 0.53 points higher than final SEER, but the unchanged rule still gains 11.79 over Full. Across three backbones, reciprocal transfer is nonnegative but sparse.

\subsection{VSR Public Random Test}

VSR \citep{liu2023vsr} provides an external binary protocol with 2,195 public random-test examples over 49 relations and 1,874 COCO images. We retain every example. To keep grounding target-relation-hidden, the grounder sees the image, subject/object phrases, and the generic question ``What is the spatial relation of [subject] to [object] in the image?''; it does not see the truth statement or relation word. Valid boxes are returned for 2,175 examples (99.09\%), and invalid cases fall back to Full. Geometry uses the same fixed $\tau=0.5$, with 97 eligible left/right or containment routes; all other relations abstain. On a reciprocal check, the entities and marker identities are swapped and the relation is replaced by its inverse, yielding a truth-equivalent statement. Hence consistency means preserving the yes/no prediction rather than inverting a relation-class prediction.

\begin{table}[!ht]
\centering
\small
\setlength{\tabcolsep}{2pt}
\begin{tabular*}{\columnwidth}{@{\extracolsep{\fill}}lrrrr@{}}
\toprule
Protocol & Accuracy & Gain & Fix/Break & $\mathrm{CI}_{img}$ \\
\midrule
Full & 77.86 & -- & -- & -- \\
Self-Marker & 78.00 & +0.14 & 131/128 & $[-1.32,+1.58]$ \\
Forward & 78.72 & +0.87 & 83/64 & $[-0.18,+1.94]$ \\
\method{} & \textbf{79.04} & +1.18 & 87/61 & $[+0.13,+2.24]$ \\
\bottomrule
\end{tabular*}
\caption{Complete VSR random-test results for Qwen3-I. Gain, Fix/Break, and the paired image-cluster 95\% interval compare each row with Full.}
\label{tab:sup_vsr}
\end{table}

Full and Marker disagree on 259 examples, but only 112 are both inverse-mappable and not already handled by threshold-eligible Geometry. Reciprocal consistency refinement is uniquely discriminative on 78 of these cases and changes 29 Forward outcomes, producing 18 fixes and 11 breaks. Its +0.32 incremental gain has image-cluster interval $[-0.14,+0.81]$ and McNemar $p=.265$; the statistically supported comparison is therefore final SEER versus Full ($p=.0395$), not SEER versus Forward. Marker helps positive statements (+1.52) but hurts negative statements ($-1.48$), illustrating why unconditional local evidence is insufficient on this heterogeneous binary benchmark.

\begin{table}[!ht]
\centering
\small
\setlength{\tabcolsep}{1pt}
\begin{tabular*}{\columnwidth}{@{\extracolsep{\fill}}llrrrr@{}}
\toprule
Split & VLM & Full & Marker & Forward & Gain \\
\midrule
Random & Qwen3-I & 77.86 & 78.00 & \textbf{78.72} & +0.87 \\
Random & Qwen3-T & \textbf{74.40} & 71.39 & 73.76 & $-0.64$ \\
Random & Qwen2.5-I & 79.09 & 75.40 & 79.00 & $-0.09$ \\
Zero-shot & Qwen3-I & \textbf{81.26} & 79.62 & 80.61 & $-0.65$ \\
\bottomrule
\end{tabular*}
\caption{Complete VSR Forward-rule transfer and split audit. Qwen3 models self-ground; Qwen2.5-I uses a fixed Qwen3-I grounder. Gain compares Forward with Full. No row-level interval excludes zero; only final Qwen3-I SEER on random test is significantly positive against Full.}
\label{tab:sup_vsr_transfer}
\end{table}

The zero-shot run was specified before scoring and reused the frozen protocol without calibration. It has no sample overlap with random test, although 261 of its 715 images are shared. Marker is unchanged on gold-true statements but loses 3.37 points on gold-false statements; reciprocal consistency refinement recovers only 0.08 points over Forward. Because truth polarity is unobservable at inference, we report this boundary rather than introduce a post-hoc polarity-conditioned rule.

The random-test model divergence is likewise polarity-sensitive. For Qwen3-T, Full already predicts yes on only 58.93\% of gold-true statements while correctly rejecting 92.41\% of gold-false statements; Marker further lowers the true-statement yes rate to 53.26\% and reduces the mean absolute margin from 0.446 to 0.336. Qwen3-I instead moves from 83.74 to 85.27 on true statements while losing 1.48 points on false statements. This pattern attributes the negative Qwen3-T transfer to marker/interface sensitivity interacting with its strong no bias, rather than to a universal failure of visual localization.

\subsection{Frozen SpatialSense Test and Protocol-Matched Diagnostic}

SpatialSense \citep{yang2019spatialsense} provides an independent binary test that was not used to design the reciprocal rule. Before model scoring, we froze all 1,860 official test annotations from six inverse predicates: left/right, above/under, and in-front-of/behind. The set contains 930 true and 930 false judgments over 1,399 images. Qwen3-I uses same-model grounding; the Qwen3-T row fixes Qwen3-I boxes so that only the answer model changes.

\begin{table}[!ht]
\centering
\small
\setlength{\tabcolsep}{1.1pt}
\begin{tabular*}{\columnwidth}{@{\extracolsep{\fill}}llrrrrrr@{}}
\toprule
VLM & $G$ & Full & Marker & Fwd. & \method{} & $\Delta$Recip. & Trig. \\
\midrule
Qwen3-I & Same & 67.69 & 67.63 & 67.53 & 67.63 & +0.11 & 265 \\
Qwen3-T & Qwen3-I & 63.82 & 62.53 & 63.66 & 63.12 & $-0.54$ & 214 \\
\bottomrule
\end{tabular*}
\caption{Frozen SpatialSense binary test (1,860 examples). $\Delta$Recip. compares final SEER with Forward; image-cluster 95\% intervals are [$-0.80$,+1.01] and [$-1.44$,+0.37].}
\label{tab:sup_spatialsense}
\end{table}

All equal-call reciprocal alternatives---Full/Marker score symmetrization, score averaging, and margin-based selection---are likewise non-significant on this frozen binary test. To distinguish a binary-polarity issue from a benchmark issue, we then constructed a protocol-matched relation-MC diagnostic from the 930 positive annotations, pairing each relation with its inverse and preserving every eligible annotation. This adaptation was created \emph{after} observing the binary result and is therefore a post-hoc mechanism diagnostic, not frozen confirmation. Qwen3-I obtains 76.67 on Full, 77.85 on Marker, 79.03 with Forward (+2.37; 56 fixes/34 breaks, McNemar $p=.0263$), and 78.71 with final SEER ($-0.32$ versus Forward). Thus the evidence interface transfers to relation classification on SpatialSense, whereas unique reciprocal consistency is not uniformly beneficial outside the protocol on which it was specified.

\section{Robustness and Boundary Analysis}

\subsection{Grounding-Order Bias}
\label{sec:sup_grounding_order_bias}

Relation hiding prevents answer leakage but does not remove prompt-order bias. Subject-first is the pre-specified realization of the ordered relation query $r(s,o)$; it is deterministic and semantically aligned with the query arguments, but is not claimed to be bias-free or empirically optimal. Hash counterbalancing instead supplies a label-independent sensitivity by assigning one of two interfaces from an arbitrary sample identifier. On 400 balanced left/right examples, subject-first grounding places the subject left of the object in 60.91\% of valid outputs, whereas object-first grounding does so in 34.26\%; independent localization is closer to balanced at 51.76\%. Direction accuracy is 65.72, 59.57, and 62.33, respectively. The bias reverses by target direction: subject-first is better for left targets, while object-first is better for right targets.

We performed a post-hoc frozen-test sensitivity on Qwen3-I and Qwen3-T without using relation labels, model outputs, or correctness for protocol assignment. Table~\ref{tab:sup_frozen_order_sensitivity} now covers complete subject-first and object-first SEER, sample-ID-hash counterbalancing, and a two-order Marker score mean. Object-first, Hash, and two-order pooling improve Full by +2.22 [0.33,4.11], +2.56 [0.72,4.39], and +3.00 [0.89,5.11]. The intervals cluster both model outcomes by image; pooled fix/break counts are descriptive. The two-order mean doubles grounding and Marker scoring but does not use Geometry or reciprocal refinement. Every pooled control is positive, showing that the aggregate conclusion is not tied to subject-first; their per-model and relation differences still establish order sensitivity. We retain subject-first as primary because it was pre-specified and uses one grounding call, not because it is assumed bias-free or universally optimal.

\begin{table*}[!tb]
\centering
\footnotesize
\setlength{\tabcolsep}{2.5pt}
\begin{tabular*}{\textwidth}{@{\extracolsep{\fill}}llrrrr@{}}
\toprule
VLM & Order & Full & \method{} & Gain [95\% CI] & F/B \\
\midrule
Qwen3-I & Subject-first & 75.33 & 78.44 & +3.11 [1.00,5.22] & 62/34 \\
Qwen3-I & Object-first & 75.33 & 76.56 & +1.22 [$-1.11$,3.56] & 63/52 \\
Qwen3-I & Hash-balanced & 75.33 & 76.56 & +1.22 [$-1.00$,3.44] & 61/50 \\
Qwen3-T & Subject-first & 70.56 & 75.33 & +4.78 [2.56,7.11] & 79/36 \\
Qwen3-T & Object-first & 70.56 & 73.78 & +3.22 [0.89,5.56] & 73/44 \\
Qwen3-T & Hash-balanced & 70.56 & 74.44 & +3.89 [1.56,6.22] & 76/41 \\
\midrule
Pooled & Subject-first & 72.94 & 76.89 & +3.94 [2.17,5.72] & 141/70 \\
Pooled & Object-first & 72.94 & 75.17 & +2.22 [0.33,4.11] & 136/96 \\
Pooled & Hash-balanced & 72.94 & 75.50 & +2.56 [0.72,4.39] & 137/91 \\
Pooled & Two-order mean & 72.94 & 75.94 & +3.00 [0.89,5.11] & 180/126 \\
\bottomrule
\end{tabular*}
\caption{Frozen GQA-Train900 grounding-order sensitivity. Hash assigns 459/441 rows to subject-/object-first by SHA256 sample-ID parity. Two-order mean averages Marker option scores across both orders. Pooled intervals resample image IDs and keep both model outcomes in one cluster.}
\label{tab:sup_frozen_order_sensitivity}
\end{table*}

The same post-hoc label-independent audit covers the four-model balanced development matrix in Table~\ref{tab:sup_balanced_order_sensitivity}. SHA256 parity assigns 450/450 GQA rows and 425/475 VG rows to subject-/object-first grounding. Full scores, candidates, option order, $\tau$, and the decision rule remain unchanged. Hash-order SEER improves all eight model--dataset rows and pools to +4.26 [3.27,5.25], slightly above the subject-first aggregate of +4.10. Seven row-wise intervals exclude zero; only R1-OV/VG remains individually unresolved. This broader audit shows that the positive cross-model aggregate is not created by a single fixed mention order, while preserving order sensitivity as a model- and relation-level limitation.

\begin{table*}[!tb]
\centering
\footnotesize
\setlength{\tabcolsep}{2.4pt}
\begin{tabular*}{\textwidth}{@{\extracolsep{\fill}}lrrrrrr@{}}
\toprule
& \multicolumn{3}{c}{GQA} & \multicolumn{3}{c}{VG} \\
\cmidrule(lr){2-4}\cmidrule(lr){5-7}
VLM & Full & Hash & Gain [95\% $\mathrm{CI}_{img}$] & Full & Hash & Gain [95\% $\mathrm{CI}_{img}$] \\
\midrule
Qwen3-I & 75.00 & 79.00 & +4.00 [1.84,6.18] & 70.33 & 73.22 & +2.89 [0.66,5.22] \\
Qwen3-T & 69.11 & 74.89 & +5.78 [3.60,7.96] & 66.67 & 72.33 & +5.67 [3.14,8.17] \\
Qwen2.5-I & 73.11 & 77.56 & +4.44 [2.38,6.51] & 68.89 & 73.89 & +5.00 [2.60,7.40] \\
R1-OV & 65.78 & 70.22 & +4.44 [2.38,6.55] & 62.67 & 64.56 & +1.89 [$-0.11$,3.92] \\
\bottomrule
\end{tabular*}
\caption{Cross-model hash-counterbalanced order audit on the relation-balanced development sets. Per-row intervals resample underlying images; the pooled interval keeps all model/dataset outcomes for an image in one cluster.}
\label{tab:sup_balanced_order_sensitivity}
\end{table*}

Pooled relation gains under Hash are +3.00 above, +3.62 behind, +5.88 below, +6.25 contains, +6.38 in-front-of, +11.50 inside, $-0.62$ left-of, $-3.25$ on, and +5.62 right-of. The image-cluster interval excludes zero for every relation except left-of; on remains negative. Thus counterbalancing removes neither semantic noise nor relation heterogeneity. Right-of is positive in this development audit despite its deficit on the frozen set, and the aggregate conclusion remains intact across answer models.

\begin{table*}[!tb]
\centering
\footnotesize
\setlength{\tabcolsep}{2.2pt}
\begin{tabular*}{\textwidth}{@{\extracolsep{\fill}}lrrrrrrrr@{}}
\toprule
& \multicolumn{3}{c}{Qwen3-I} & \multicolumn{3}{c}{Qwen3-T} & \multicolumn{2}{c}{Pooled} \\
\cmidrule(lr){2-4}\cmidrule(lr){5-7}\cmidrule(lr){8-9}
Relation & Full$\to$SEER & Gain [CI] & $p$ & Full$\to$SEER & Gain [CI] & $p$ & Full$\to$SEER & Gain [cluster CI] \\
\midrule
left-of & 58$\to$67 & +9 [$0$,18] & .093 & 60$\to$68 & +8 [$-2$,18] & .169 & 59$\to$67.5 & +8.5 [0.5,16.5] \\
right-of & 81$\to$77 & $-4$ [$-11$,3] & .424 & 69$\to$68 & $-1$ [$-9$,7] & 1.00 & 75$\to$72.5 & $-2.5$ [$-8.5$,3] \\
above & 61$\to$64 & +3 [$-3$,9] & .508 & 56$\to$59 & +3 [$-1$,8] & .375 & 58.5$\to$61.5 & +3 [$-1$,7.5] \\
below & 64$\to$67 & +3 [$-3$,9] & .549 & 65$\to$68 & +3 [$-2$,8] & .453 & 64.5$\to$67.5 & +3 [$-1$,7.5] \\
on & 80$\to$72 & $-8$ [$-15$,$-2$] & .039 & 77$\to$77 & 0 [$-5$,5] & 1.00 & 78.5$\to$74.5 & $-4$ [$-9$,0.5] \\
front & 74$\to$84 & +10 [3,17] & .013 & 68$\to$74 & +6 [$-1$,13] & .180 & 71$\to$79 & +8 [2.5,14] \\
behind & 86$\to$81 & $-5$ [$-10$,0] & .125 & 85$\to$81 & $-4$ [$-8$,$-1$] & .125 & 85.5$\to$81 & $-4.5$ [$-8.5$,$-1$] \\
inside & 82$\to$89 & +7 [2,13] & .039 & 64$\to$88 & +24 [16,33] & $<.001$ & 73$\to$88.5 & +15.5 [10,21] \\
contains & 92$\to$88 & $-4$ [$-11$,3] & .388 & 91$\to$87 & $-4$ [$-11$,3] & .388 & 91.5$\to$87.5 & $-4$ [$-9.5$,1.5] \\
\bottomrule
\end{tabular*}
\caption{Per-relation hash-counterbalanced results on the frozen set (100 examples per relation and model). Intervals are paired image bootstrap intervals; pooled intervals cluster the two model outcomes by image. Per-model $p$ values are exact McNemar tests and are exploratory without correction across nine relations.}
\label{tab:sup_frozen_order_relations}
\end{table*}

\FloatBarrier

We also tested whether multiple localization orders could become a stronger training-free route on the earlier 400-example horizontal diagnostic. The original full-image accuracy is 65.00; subject-first, object-first, and independent marker states achieve 62.00, 62.50, and 57.25. Selecting the maximum-gap state among Full and the joint subject-first Marker reaches 65.25, while selecting among all grounding orders reaches 64.00. An oracle over grounding views reaches 75.75, showing headroom, but the tested observable rules do not realize it. We therefore retain one target-relation-hidden grounding call and report horizontal bias as a limitation rather than adding costly multi-order inference.

\begin{table}[!ht]
\centering
\small
\setlength{\tabcolsep}{2pt}
\begin{tabular*}{\columnwidth}{@{\extracolsep{\fill}}lrrr@{}}
\toprule
Evaluation slice & N & Full & \method{} / gain \\
\midrule
Development: all relations & 41,544 & 70.60 & 74.60 / +3.99 \\
\quad without left/right & 34,140 & 72.26 & 76.34 / +4.09 \\
\quad left/right only & 7,404 & 62.99 & 66.55 / +3.55 \\
\quad worst final relation: on & 5,380 & 77.99 & 73.94 / $-4.05$ \\
\midrule
Frozen: all relations & 1,800 & 72.94 & 76.89 / +3.94 \\
\quad without left/right & 1,400 & 74.64 & 78.21 / +3.57 \\
\quad left/right only & 400 & 67.00 & 72.25 / +5.25 \\
\quad without inside/contains & 1,400 & 70.29 & 72.93 / +2.64 \\
\quad without target on & 1,600 & 72.25 & 77.00 / +4.75 \\
\quad without role-swapped rows & 1,600 & 70.62 & 74.75 / +4.13 \\
\bottomrule
\end{tabular*}
\caption{Relation and construction robustness for final SEER. Frozen rows pool Qwen3-I/T on GQA-Train900. Gains remain positive after removing horizontal, containment, or role-swapped constructions. Reciprocal consistency refinement raises development-pool right-of accuracy from 57.67 under Forward to 59.78, reducing its deficit against Full from $-5.89$ to $-3.78$; the worst final relation is instead on.}
\label{tab:sup_direction_robustness}
\end{table}

\subsection{Text-Only and Shuffled-Image Controls}

We freeze a balanced Qwen3-I control set containing 1,800 forward examples (GQA900+VG900) and all 1,200 exact-inverse reciprocal queries. Text-only retains the entity names, candidate options, and Full- or Marker-style instruction but removes the image. Image-shuffle assigns every forward example an image from a different source example and gives its reciprocal query the same incorrect image. Geometry is disabled so that all decisions compare Full and Marker interfaces. Table~\ref{tab:sup_visual_destruction} shows a 10--13 point intact-evidence advantage, confirming that the task is not solved by text alone. However, reciprocal selection still gains after visual destruction and image-shuffle produces many uniquely consistent cases. Reciprocal consistency is therefore an observable transformation cue between states, not a sufficient certificate of visual correctness; final SEER consequently uses it only on inverse-supported Full/Marker conflicts and otherwise retains Forward.

\begin{table}[!ht]
\centering
\footnotesize
\setlength{\tabcolsep}{0.4pt}
\begin{tabular*}{\columnwidth}{@{\extracolsep{\fill}}lrrrrrrr@{}}
\toprule
Condition & Full & Mark. & Fwd. & Rec. & $\Delta$ & F/B & U\% \\
\midrule
Intact visual & 72.67 & 74.39 & 74.11 & 75.50 & +1.39 & 47/22 & 27.08 \\
Text-only & 62.39 & 63.89 & 63.06 & 64.39 & +1.33 & 46/22 & 20.42 \\
Image-shuffle & 60.22 & 62.28 & 62.94 & 65.33 & +2.39 & 73/30 & 41.42 \\
\bottomrule
\end{tabular*}
\caption{Visual-destruction controls. Accuracy and reciprocal-versus-Forward Fix/Break are on 1,800 examples; ``Unique'' is the percentage of 1,200 inverse-supported examples for which exactly one state is reciprocal-consistent. Similar positive paired changes without correct images show that reciprocity measures interface transformation consistency, not visual correctness. No Geometry route is used.}
\label{tab:sup_visual_destruction}
\end{table}

The matched sample identities show that the three gains do not come from repeatedly changing the same language-favored examples. Of 69 intact-image answer changes, only 3 overlap with text-only changes and 7 with shuffled-image changes; all 3 and 5 of the 7, respectively, have the same answer transition. Only 2/47 intact fixes remain fixes under text-only input and 3/47 under image shuffle. Trigger-set Jaccard similarities are .138/.209, while actual answer-change Jaccards are only .022/.042. Thus visual destruction can create its own protocol-consistent changes, but these are largely different from the intact-image corrections. Stratifying intact rows by the minimum source-box IoU gives net gains of +1.16, +1.58, and +2.04 for IoU below .25, .25--.50, and at least .50; the cue is not restricted to high-IoU grounding, though breaks become rarer there.

\subsection{HallusionBench Interface Diagnostic}

HallusionBench \citep{guan2024hallusionbench} exposes answer-interface sensitivity in reasoning models. Under an 8-token direct-generation budget, R1-OV and Kimi-VL begin with reasoning text and yield no parseable yes/no answer. Closed-set yes/no log-probability removes this formatting artifact. This diagnostic motivates controlled scoring in the main experiments; it is not evidence that closed-set scoring is universally preferable.

\begin{table}[!ht]
\centering
\small
\setlength{\tabcolsep}{2pt}
\begin{tabular*}{\columnwidth}{@{\extracolsep{\fill}}llrr@{}}
\toprule
Model & Protocol & Parse & Accuracy \\
\midrule
Qwen3-T & Direct, 8 tok. & 99.50 & 68.50 \\
Qwen3-I & Direct, 8 tok. & 100.00 & 63.00 \\
Qwen2.5-I & Direct, 8 tok. & 100.00 & 66.50 \\
R1-OV & Direct, 8 tok. & 0.00 & 0.00 \\
Kimi-VL & Direct, 8 tok. & 0.00 & 0.00 \\
R1-OV & Yes/no log-prob. & 100.00 & 69.00 \\
Kimi-VL & Yes/no log-prob. & 100.00 & 54.00 \\
\bottomrule
\end{tabular*}
\caption{HallusionBench 200-example format diagnostic. Unparsed direct outputs count as wrong.}
\label{tab:sup_hallusion}
\end{table}

\subsection{Non-Spatial Boundary}

On a 1,651-example SugarCrepe \citep{hsieh2023sugarcrepe} caption subset, Qwen3-T already scores 92.49 on the full image. A caption-difference crop reaches 92.55 (+0.06), while adding boxes reduces accuracy to 91.22 (-1.27). This boundary supports applying explicit evidence states when a decision genuinely requires subject/object localization, not as an unconditional preprocessing step for all compositional tasks.

\FloatBarrier

\section{Reproducibility Notes}

\noindent\textbf{Hardware, software, and decoding.} The reported Qwen-family runs use one NVIDIA A800 80GB PCIe GPU under CentOS 7, Python 3.13.9, PyTorch 2.6.0+cu124, Transformers 4.57.1, Accelerate 1.11.0, and qwen-vl-utils 0.0.14. Inference uses bfloat16 and batch size one. Grounding is greedy with temperature zero and top-$p$ disabled, using at most 96 new tokens for the main GQA/VG protocols; closed-set answers are length-normalized continuation log-probabilities scored sequentially. The processor preserves aspect ratio with \texttt{min\_pixels=max\_pixels=50,176}.

\begin{table}[!ht]
\centering
\small
\setlength{\tabcolsep}{2pt}
\begin{tabular*}{\columnwidth}{@{\extracolsep{\fill}}lll@{}}
\toprule
Model & Checkpoint ID & Rev. prefix \\
\midrule
Qwen3-I & \texttt{Qwen3-VL-8B-Instruct} & \texttt{0c351dd01ed8} \\
Qwen3-T & \texttt{Qwen3-VL-8B-Thinking} & \texttt{92f3c4b4fead} \\
Qwen2.5-I & \texttt{Qwen2.5-VL-7B-Instruct} & \texttt{cc594898137f} \\
R1-OV & \texttt{R1-Onevision-7B} & \texttt{44ec19bbe90f} \\
InternVL3.5 & \texttt{InternVL3\_5-8B-HF} & \texttt{741a7d030204} \\
\bottomrule
\end{tabular*}
\caption{Checkpoint revision prefixes used in the reported evaluations.}
\label{tab:sup_model_revisions}
\end{table}

\noindent\textbf{Protocol chronology.} The evidence states, Forward rule, and reciprocal unique-consistency rule were developed on the GQA/VG validation diagnostics. Before the first Qwen3-I EmbSpatial evaluation, we fixed $\tau=0.5$, invalid-grounding fallback, reciprocal eligibility, and the unique-consistency decision. Subsequent Qwen3-T and InternVL3.5 EmbSpatial replications used these settings unchanged. After method development and before any scoring on GQA-Train900, we froze its image-disjoint IDs and checksum; this set therefore confirms final SEER rather than selecting it. The grounding-order and No-Geometry audits were designed afterward; they reuse fixed examples, state scores, and decision rules and are reported as robustness or mechanism analyses, not protocol confirmation. The complete 1,860-example SpatialSense inverse-predicate binary test was likewise frozen before scoring. Its null/negative result bounds transfer from inverse-supported relation choice to binary truth judgment; the 930-example SpatialSense relation-MC adaptation was constructed only afterward and is labeled post hoc. Human labels were collected after the corresponding predictions were fixed and were used only for the audits in Tables~\ref{tab:sup_human_audit}, \ref{tab:sup_human_clean}, \ref{tab:sup_frozen_human_strata}, and \ref{tab:sup_frozen_human_clean}, never for routing or threshold selection. Tables~\ref{tab:sup_embspatial}, \ref{tab:sup_vsr_transfer}, and \ref{tab:sup_spatialsense} report external transfer and boundaries; no model or decision component is fit to their labels.

\noindent\textbf{Possible pretraining exposure.} Image-disjoint construction and protocol freezing prevent overlap with our development pools and prohibit selection using test outputs; they do not reveal the proprietary pretraining corpora of evaluated checkpoints. We therefore cannot certify that GQA/VG images or annotations were absent from pretraining. The causal comparisons are paired within model and item, so Full and SEER share any memorized content while differing only in the evidence interface. Architecture-diverse replications, EmbSpatial transfer, and image-destruction controls reduce reliance on absolute benchmark accuracy, but they cannot eliminate a possible interaction between prior exposure and the intervention.

All comparisons preserve identical sample identifiers, candidates, targets, and option order across Full and Marker. Reused Qwen3-I predictions were required to satisfy exact input equivalence; a 494-example audit confirmed identical predictions and zero maximum continuation-NLL difference. Target-relation-hidden boxes required the same match, while role-swapped containment examples were grounded anew. Per-example analyses use predictions, routes, margins, Geometry layout scores, and paired correction counts.

Remaining nondeterminism comes from model kernels and generation; dataset construction, candidate order, parsing, geometry, and evidence decisions are deterministic under seed 2027. Image-cluster intervals use 20,000 paired bootstrap repetitions. Because GQA is derived from VG, identical numeric image IDs are merged across both pools, and all VLM outcomes for one underlying image share a cluster. The method trains no component. Forward uses one Full score, one target-relation-hidden grounding generation, and one Marker score; Geometry and decision logic add no VLM call. Final SEER adds two state scores only on eligible reciprocal conflicts.

\end{document}